\documentclass{dukecei}

\usepackage[utf8]{inputenc}
\usepackage{microtype}
\usepackage{url}
\usepackage{placeins}
\usepackage{flafter}

\makeatletter
\let\algorithmic\@undefined
\let\endalgorithmic\@undefined
\@for\@cei@tmp:={require,ensure,end,if,then,else,elsif,endif,for,forall,do,%
  endfor,while,endwhile,loop,endloop,repeat,until,print,return,comment,true,%
  false}\do{\expandafter\let\csname algorithmic\@cei@tmp\endcsname\@undefined}
\makeatother
\usepackage{algpseudocode}

\theoremstyle{definition}
\newtheorem{ceidefinition}[definition]{Definition}


\title{What Does a Token Cost? A Mixture-of-Agents Measurement of Sufficient Per-Token Compute}

\author{
    Zhixu Du$^{1}$\footnote{Correspondence E-mail: [zhixu.du@duke.edu]},
    Weijia Han$^{2}$,
    Hai Helen Li$^{1}$, 
    Yiran Chen$^{1}$\\[1em]
    \normalsize $^{1}$Duke University \\
    \normalsize $^{2}$University of Washington
}

\begin{document}

\maketitle
\thispagestyle{firstpagestyle} 

\begin{abstract}
Large language models spend the same amount of computation on every token they
generate, regardless of how difficult each token is to produce. Methods such
as speculative decoding and model routing are built on the premise that much
of this computation is unnecessary, yet the computation an individual token
actually requires has not been measured. We measure it through a
Mixture-of-Agents (MoA) lens: a panel of fifteen language models of
increasing capacity, drawn from three families, in which every
agent attempts to reproduce a reference sequence token by token, conditioned
on the correct preceding tokens. We define the inference cost of the smallest
agent that succeeds as the token's sufficient compute, which upper-bounds
what the token requires. On three core benchmarks, a 0.5B agent reproduces 92--95\% of reference tokens. Across Qwen, OLMo, and R1-distilled panels, the most expensive 10\% account for 64--80\% of estimated FLOPs. 
On all 500 MATH-500 problems, the MoA-derived map helps model routing reduce projected latency from 7.59 to 5.12 seconds while slightly improving accuracy, relative to the best confidence-routing baseline. The MoA-map helps drafting use 32.6\% fewer draft tokens and
approximately 20\% lower projected latency than fixed-window drafting at similar accuracy. These  comparisons reveal remaining allocation headroom, motivating controllers that exploit sufficient-compute structure.
\end{abstract}

\section{Introduction}
\label{sec:intro}

Large language models (LLMs), such as the GPT series, have made a great impact for tasks ranging from writing to programming
\citep{brown2020fewshot,openai2023gpt4}. Their adoption produces
recurring computational expense at a large scale: by July 2025, ChatGPT alone
received more than 2.5 billion user messages per day
\citep{chatterji2025usage}. Therefore, efficiently serving these requests is becoming critical \citep{pope2023inference,kwon2023vllm}.
One approach to reducing this computational cost is to use only as much computation as is necessary. \textit{Speculative decoding} lets a small model
propose several tokens for a larger model to verify together
\citep{leviathan2023speculative,chen2023accelerating}.\textit{ Model cascades and routers}
choose among models according to the input
\citep{chen2023frugalgpt,ong2025routellm}, while collaborative decoding makes that
choice within a response \citep{shen2024colllm,fu2025r2r}. \textit{Adaptive-depth} methods
similarly vary the layers evaluated for each prediction
\citep{schuster2022calm,raposo2024mod}. These approaches exploit the possibility
that a cheaper computation can supply a useful prediction. Their gains depend on how often this is possible. However, running the model alone does not reveal when less computation would have been sufficient.

\begin{figure}[t]
\centering
\includegraphics[width=\textwidth]{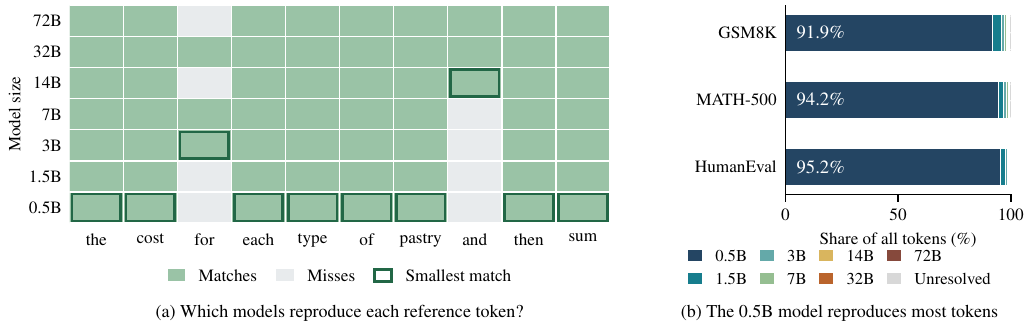}
\caption{Measuring which model suffices at each token.
(a) Each column is a token from a GSM8K response. Every model
receives the same reference prefix; green cells reproduce the next token and
outlines mark the smallest matching model.
(b) Within the Qwen2.5 family, the 0.5B model reproduces 91.9--95.2\%
of tokens. The ``unresolved''
means that none matches.}
\label{fig:overview}
\end{figure}

This motivates our central question: \emph{How much computation is sufficient
for each token in a successful response?} A response-level accuracy or cost average hides where that computation is needed. Even a high agreement rate between a small and a large model leaves
open how much computation the remaining disagreements require. Existing
measurements of oracle module allocation, small/large-model disagreement, and
learned layer use establish that computation can vary within a response
\citep{alizadeh2024duollm,zhang2024fsgen,luo2025flexidepth}. We measure this
variation by identifying, for each token, the smallest model that can successfully reproduce it. We then examine whether this token-level information can improve existing efficient serving methods.

Mixture-of-Experts
(MoE) has been widely known, which selects subnetworks inside a model
\citep{shazeer2017moe,fedus2022switch}. Mixture-of-Agents (MoA)combines different language models, either through response aggregation \citep{wang2024moa} or through
agent selection during decoding \citep{chakraborty2025collab,zheng2025citer}. Our measurement uses token-level MoA. Given a reference sequence, we query a panel
of fifteen models from Qwen2.5, OLMo-2, and DeepSeek-R1-Distill-Qwen at each token.
The FLOPs of the cheapest agent whose prediction matches the next token
are that token's \emph{sufficient compute}. A successful agent supplies an
observed upper bound on the computation cost of that token.

Figure~\ref{fig:overview} provides an overview of our measurement. Within the Qwen2.5 panel, the 0.5B agent reproduces 91.9--95.2\% of reference tokens on GSM8K, MATH-500, and HumanEval. The remaining tokens are few but expensive. In all three panels, the most expensive 10\% of tokens account for 64--80\% of estimated FLOPs. Families also agree on which positions are expensive. After controlling for surprisal and position, the reproduction orderings of Qwen2.5, OLMo-2, and Falcon3 families have pairwise partial rank correlations of 0.32--0.48. This agreement makes the map informative beyond the behavior of a single model family. Details will be discussed in Section~\ref{sec:agreement}.

We then ask whether this map would help existing methods allocate computation. On all 500 MATH-500 problems, we give a routing controller and a speculative-decoding controller oracle access to the map of 0.5B agreement along a 72B reference response. Routing to 0.5B at positions the map marks as reproducible reaches 69.6\% accuracy at a projected latency of 5.12 seconds, compared with 68.4\% and 7.59 seconds for the best confidence-threshold cascade. Setting each draft length to the map's run of reproducible positions proposes 32.6\% fewer draft tokens than a fixed 16-token window and lowers projected latency by about 20\% at similar accuracy. 
Our contributions can be summarized as:

\begin{enumerate}
  \item We formalize sufficient compute through token-level MoA and introduce a panel-based method for measuring it along verified responses.
  \item We show that a sparse tail of tokens carries most of the estimated compute and that three families share a reproduction ordering. 
  \item We evaluate how routing and speculative decoding exploit the measurement. Both improve projected efficiency at comparable accuracy over baselines. The difference is allocation headroom that current controllers leave unrealized.
\end{enumerate}

 In this paper, we first define sufficient compute
  (Section~\ref{sec:formulation}) and derive a token-level MoA procedure
  for measuring it (Section~\ref{sec:instrument}). The experiments then
  examine where compute is concentrated, whether cheap reproduction supports
  continued generation, and how existing efficiency methods exploit this
  information (Section~\ref{sec:experiments}). Section~\ref{sec:related} places our study in the context of prior work.
  Section~\ref{sec:discussion} discusses the implications and limitations
  of our findings, and Section~\ref{sec:conclusion} concludes the paper.
\section{Problem formulation and Background}
\label{sec:formulation}
\textbf{Sufficient compute at a reference prefix}
\label{sec:sufficient-definition}
Let $x$ be a prompt and $y^\star=(y_1^\star,\ldots,y_T^\star)$ a reference
accepted by a binary task checker $V(x,y)\in\{0,1\}$, so $V(x,y^\star)=1$. At position $t$, the
reference state is $s_t^\star=(x,y_{<t}^\star)$: the prompt and all preceding
reference tokens. A model panel
$\Pi_F=\{\pi_{F,1},\ldots,\pi_{F,K_F}\}$ contains next-token distributions
$\pi_{F,k}(v\mid s)$ over a shared vocabulary $\mathcal V$.
The index $F$ identifies the panel. Let $c_{F,k}(s)>0$ be the FLOPs of model
$k$'s next-token prediction at prefix $s$.
A prediction-success rule $M(\pi,s,v)\in\{0,1\}$ specifies when a model
succeeds on target token $v$ at state $s$. We require the
model's prediction to equal the reference token.

\begin{definition}[Panel-relative sufficient compute]
\label{def:sufficient}
For reference token $y_t^\star$, define the successful agents and their
minimum prediction cost by
\begin{equation}
 \mathcal A_F^M(t)=\{k:M(\pi_{F,k},s_t^\star,y_t^\star)=1\},\qquad
 d_F^M(t)=\min_{k\in\mathcal A_F^M(t)}c_{F,k}(s_t^\star).
 \label{eq:suf}
\end{equation}
We set $\min\varnothing=+\infty$ and call such a position \emph{unresolved}. Appendix~\ref{app:concentration-conventions} reports how often this occurs and how such positions enter the cost summaries.
The sequence $d_F^M(1{:}T)$ is the sequence's sufficient-compute map.
We write $d_F(t)$ when the success rule is fixed. 
\end{definition}
It serves as an upper bound since adding agents can only tighten the bound. The measured map depends on the reference
trajectory, and another correct response may have a different cost profile.

\textbf{Speculative decoding and Routing}
The map makes precise a local opportunity shared by several efficiency methods:
a prediction produced by a large model may also be available from a cheaper
one. Speculative decoding uses a draft model to propose multiple tokens and
verifies them with a target model in parallel
\citep{leviathan2023speculative,chen2023accelerating}. Its distribution-preserving
acceptance rule depends on draft/target probabilities, and its speed benefit
depends on accepted tokens per verification round and the relative cost of
drafting and verification. Greedy reference agreement gives a concrete local
comparison; speculative sampling uses the full distributions to decide acceptance.

Routing changes which model supplies the output. FrugalGPT and RouteLLM select
models at the request level \citep{chen2023frugalgpt,ong2025routellm}; collaborative
decoding extends selection to individual steps
\citep{shen2024colllm,zheng2025citer,fu2025r2r}. At that granularity, a cheaper
prediction also changes the prefix on which later choices are made. A cost map
measured at reference states consequently supplies information about those
states, while a deployed controller must act along the states it generates.
This distinction motivates both the exact-reference selector and the later
continuation experiments.

\textbf{From MoE to token-level MoA}
Mixtures provide a way to express the choice of computation. An MoE layer
maps a token representation $h$ to $h'$ by combining $J$ expert functions
$E_j$ through a gate $w_j$:
\begin{equation}
 h'=\sum_{j=1}^{J}w_j(h)E_j(h),\qquad
 w_j(h)\ge0,\quad\sum_{j=1}^{J}w_j(h)=1.
 \label{eq:moe}
\end{equation}
The experts are subnetworks within a model \citep{jacobs1991mixtures}; sparse
gates activate only a subset at each token
\citep{shazeer2017moe,fedus2022switch}. This separates parameter capacity from
the computation performed for a particular prediction.

MoA makes language models the available policies. In response-level
MoA, agents produce candidate answers and an aggregator conditions on them,
possibly across several rounds \citep{wang2024moa}. Token-level MoA selects
an agent's next-token policy at a shared decoding state
\citep{shen2024colllm,chakraborty2025collab,zheng2025citer}. For our measurement,
this gives a direct interpretation of $\mathcal A_F^M(t)$: it is the set of
agents that can supply the required prediction at the current reference state.
Section~\ref{sec:instrument} derives the controller that selects among those
agents.

\section{Measuring sufficient compute with token-level MoA}
\label{sec:instrument}

\begin{figure}[t]
    \centering
    \includegraphics[width=\textwidth]{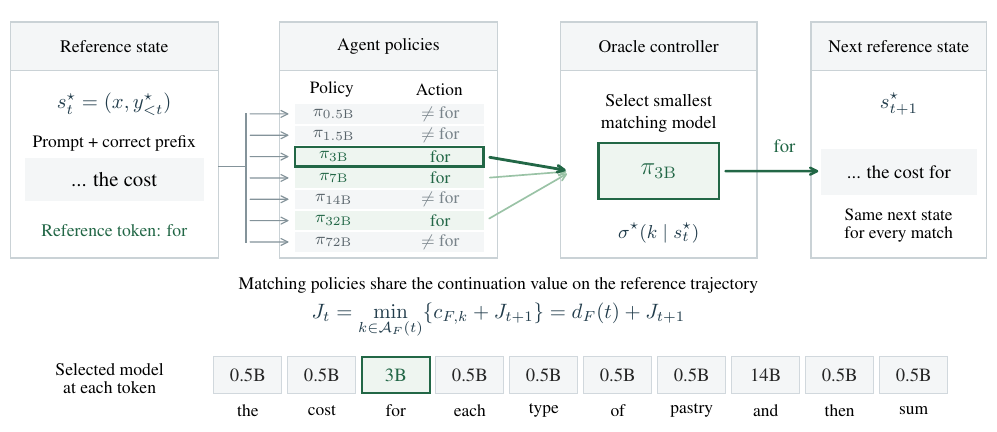}
    \caption{
    Minimum-cost policy selection.
    At the illustrated prefix, the 3B, 7B, and 32B agents all reproduce
    the reference token ``for'' and reach the same next state.
    Their remaining cost $J_{t+1}$ is therefore identical, so the
    controller selects the agent with the lowest immediate prediction
    FLOPs: 3B in this example.
    Repeating this selection gives the model sequence shown along the
    bottom.
    }
    \label{fig:moa}
\end{figure}

To connect sufficient compute to sequence generation, we treat each model
as a policy and model selection as a decision made at every token. We first
formulate this MoA process, then derive the minimum-cost controller for
reproducing a verified response.

\textbf{Token-level MoA as a Markov Decision Process.}
\label{sec:moa-mdp}
Decoding can be represented as a token-level Markov decision process
\citep{ranzato2016sequence,bahdanau2017actor,mudgal2024controlled,
chakraborty2024transfer}. At step $t$, the state
$s_t=(x,y_{<t})$ contains the prompt and the tokens generated so far.
The action is the next token $v$. Appending this token deterministically
produces the next state, $s_{t+1}=(x,y_{<t}v)$.
The state contains all information used by the next-token policies,
satisfying the Markov property. Generation terminates upon emitting an
end-of-sequence token or reaching a prescribed maximum length.
The resulting response $y$ constitutes a \emph{rollout}, evaluated by
the terminal reward $V(x,y)$.

In token-level MoA, a controller chooses which model supplies the
next-token policy \citep{chakraborty2025collab}.
Each agent $\pi_{F,k}\in\Pi_F$ defines a distribution over the shared
vocabulary. Let $\sigma(k\mid s)$ denote the probability of selecting
agent $k$ at state $s$, where $\sigma(k\mid s)\geq 0$ and
$\sum_{k=1}^{K_F}\sigma(k\mid s)=1$.
The controller first selects an agent and then samples a token from
that agent's policy. Marginalizing over the selected agent gives
\begin{equation}
    \pi_\sigma(v\mid s)
    =
    \sum_{k=1}^{K_F}
    \sigma(k\mid s)\pi_{F,k}(v\mid s).
    \label{eq:moa}
\end{equation}
Thus, the controller selects among model policies, while the selected
policy determines the next token. Different agents can produce different
tokens and therefore lead to different subsequent states.

We measure prediction cost using the FLOP functions $c_{F,k}(s)$
defined in Section~\ref{sec:formulation}. We consider an idealized
setting in which each agent's prefix cache is available and switching
between agents incurs no additional cost.
Let $K_t$ denote the agent selected at step $t$, and let $T(Y)$ be
the length of the generated response $Y$.
For an expected compute budget $B$, the controller's objective is
\begin{equation}
    \max_\sigma\;
    \mathbb{E}_\sigma[V(x,Y)]
    \quad\text{subject to}\quad
    \mathbb{E}_\sigma
    \left[
        \sum_{t=1}^{T(Y)} c_{F,K_t}(s_t)
    \right]
    \leq B.
    \label{eq:budget-objective}
\end{equation}
The expectation accounts for both agent selection and token generation.
An agent choice affects its immediate prediction cost and the continuation
from the resulting prefix.

\begin{algorithm}[t]
\caption{Measuring sufficient compute by token-level policy selection}
\label{alg:measurement}
\begin{algorithmic}
    \Require Prompt $x$; verified reference $y^\star_{1:T}$;
    panel $\Pi_F$ with FLOP functions $c_{F,k}(s)$
    \Ensure Matches $A_F$; selected agents $k^\star_{1:T}$;
    costs $d_F(1{:}T)$; unresolved positions $U$
\end{algorithmic}
\smallskip
\begin{minipage}[t]{0.48\linewidth}
\centering Evaluate reference states\par
\begin{algorithmic}[1]
    \State $U\gets\varnothing$
    \For{$t=1,\ldots,T$}
        \State $s_t^\star\gets(x,y_{<t}^\star)$
        \For{$k=1,\ldots,K_F$}
            \State $a_{F,k,t}\gets
                \mathbf{1}[g_{F,k}(s_t^\star)=y_t^\star]$
            \State $c_{F,k,t}\gets c_{F,k}(s_t^\star)$
        \EndFor
        \State $\mathcal A_F(t)\gets\{k:a_{F,k,t}=1\}$
        \algstore{measurement}
\end{algorithmic}
\end{minipage}\hfill
\begin{minipage}[t]{0.48\linewidth}
\centering Select least-cost agents\par
\begin{algorithmic}[1]
        \algrestore{measurement}
        \If{$\mathcal A_F(t)\neq\varnothing$}
            \State $k_t^\star\gets
                \arg\min_{k\in\mathcal A_F(t)}c_{F,k,t}$
            \State $d_F(t)\gets c_{F,k_t^\star,t}$
        \Else
            \State $k_t^\star\gets\bot$;
                $d_F(t)\gets+\infty$
            \State $U\gets U\cup\{t\}$
        \EndIf
    \EndFor
    \State \Return $A_F$, $k^\star_{1:T}$, $d_F(1{:}T)$, $U$
\end{algorithmic}
\end{minipage}
\end{algorithm}

\textbf{Exact-reference selection and measurement.}
To measure sufficient compute, we fix a verified response $y^\star$
and ask which agents can reproduce it at minimum cost.
Define
\begin{equation}
    g_{F,k}(s)
    =
    \arg\max_{v\in\mathcal V}\pi_{F,k}(v\mid s),
    \qquad
    a_{F,k,t}
    =
    \mathbf{1}
    \left[g_{F,k}(s_t^\star)=y_t^\star\right].
    \label{eq:measurement}
\end{equation}
The matrix $A_F=(a_{F,k,t})$ records which agents reproduce which
reference tokens. At position $t$, the admissible agents are
$\mathcal A_F(t)=\{k:a_{F,k,t}=1\}$, and their prediction costs are
$c_{F,k,t}=c_{F,k}(s_t^\star)$.
If every reference position is resolved, then
$\mathcal A_F(t)\neq\varnothing$ for all $t$.
A controller reproduces the reference by choosing an admissible agent
at each position. The minimum-cost selection problem is therefore
\begin{equation}
    \min_{(k_1,\ldots,k_T)\in
          \prod_{t=1}^{T}\mathcal A_F(t)}
    \sum_{t=1}^{T}c_{F,k_t,t}.
    \label{eq:reference-objective}
\end{equation}

The key simplification is that every admissible agent emits the same
token $y_t^\star$ and reaches the same next state $s_{t+1}^\star$.
Consequently, choosing among these agents changes the immediate cost
without changing the remaining reference or the available future choices.
Let $J_t$ be the minimum cost of reproducing the remaining reference
from $s_t^\star$. Under the additive-cost assumptions above, its
Bellman recursion becomes
\begin{equation}
    J_t
    =
    \min_{k\in\mathcal A_F(t)}
    \left(c_{F,k,t}+J_{t+1}\right)
    =
    d_F(t)+J_{t+1},
    \qquad
    J_{T+1}=0.
    \label{eq:bellman}
\end{equation}
The continuation cost $J_{t+1}$ is shared by all admissible agents.
Minimizing the total reference cost therefore reduces to selecting
the least costly matching agent at each position:
\begin{equation}
    k_t^\star
    =
    \arg\min_{k\in\mathcal A_F(t)}c_{F,k,t},
    \qquad
    \sigma_{y^\star}^\star(k\mid s_t^\star)
    =
    \mathbf{1}[k=k_t^\star].
    \label{eq:minimum-selector}
\end{equation}
This selector reproduces the reference from the initial prompt,
with minimum cost: 
\begin{equation}
    C_{F,\mathrm{ref}}^\star(x,y^\star)
    =
    J_1
    =
    \sum_{t=1}^{T}d_F(t).
    \label{eq:selector}
\end{equation}
Figure~\ref{fig:moa} illustrates this tokenwise measurement.
Because $V(x,y^\star)=1$, the selected agent sequence supplies a
correct response at cost $C_{F,\mathrm{ref}}^\star(x,y^\star)$.
It upper-bounds the minimum cost of generating any correct
response
\(
    C_{F,\mathrm{correct}}^\star(x)
    \leq
    C_{F,\mathrm{ref}}^\star(x,y^\star).
\)
Appendix~\ref{app:bound} specifies the candidate class, states the assumptions under which Eq.~\eqref{eq:selector} is the minimum over all selectors that reproduce $y^\star$, and gives the local witness bound for larger model classes.

Algorithm~\ref{alg:measurement} computes the matching matrix and the
resulting selector. 
If no agent matches a token, the algorithm records that position as
unresolved and assigns $d_F(t)=+\infty$.
We implement the reference-state evaluations using teacher forcing:
a causal forward pass supplies each model with only the preceding
reference tokens at each prediction position.
We retain both the selected agent identity $k_t^\star$ and its
prediction FLOPs $d_F(t)$.
The reported sufficient compute measures the selected agent's
prediction cost.
The resulting measurements characterize compute along verified
trajectories.

\section{Experiments}
\label{sec:experiments}

\label{sec:setup}
\textbf{Models.}
  Our main measurement panel comprises fifteen models across three families:
  Qwen2.5 (0.5B, 1.5B, 3B, 7B, 14B, 32B, 72B) \citep{qwen2024qwen25}, OLMo-2 (1B, 7B, 13B, 32B) \citep{olmo2025olmo2}, and DeepSeek-R1-Distill-Qwen (1.5B, 7B, 14B, 32B)
  \citep{guo2025deepseek} as
  reasoning-distilled candidates. Falcon3 \citep{falcon2024falcon3} (1B, 3B, 7B, 10B) is tested as extended family candidates.
  The reference generator samples four responses per prompt at temperature 0.7
  and top-$p$ 0.95. Qwen2.5-72B-Instruct generates the references used throughout; Llama-3.3-70B-Instruct generates a second reference set for the cross-family analysis. Table~\ref{tab:models} lists every model role, including a Qwen3-30B-A3B MoE generator as a sensitivity check (Appendix~\ref{app:controls}).
  Llama-3.1-8B-Instruct \citep{grattafiori2024llama3} supplies token
  surprisal for statistical controls.
  Appendix~\ref{app:data} provides additional setup details.

\textbf{Datasets.}
The main experiments use three core datasets covering mathematical reasoning
and code generation: GSM8K (\citealp{cobbe2021gsm8k}),
MATH-500 (\citealp{lightman2023verify}), and HumanEval
(\citealp{chen2021humaneval}).
Nine additional benchmarks extend the analysis to competition mathematics,
programming, question answering, knowledge, and commonsense reasoning,
including LiveCodeBench \citep{jain2024livecodebench}, BBH
\citep{suzgun2022bbh}, and MMLU-Pro \citep{wang2024mmlupro}.
Together, the twelve benchmarks contain 50,019 evaluation prompts.
Table~\ref{tab:datasets} lists all datasets.

\textbf{Hardware and software.}
Generation and teacher-forced scoring run on a server equipped with
eight 80\,GB NVIDIA A100-SXM4 GPUs.
We use vLLM 0.14.1 \citep{kwon2023vllm} with BF16 precision.

\textbf{Measurement.}
Our primary criterion is exact reproduction of the reference token. Appendix~\ref{app:methods} defines the teacher-forced scoring and the top-five and $p \geq 0.5$ alternatives, whose evaluation results are in Table~\ref{tab:robustness}.
We record the selected model and its architecture-derived FLOPs.
The count includes attention, projection, and feed-forward operations.
To compare families with different
tokenizers, we align their predictions to common words and assign each word
the maximum selected model index among its tokens; unresolved words receive
an ordinal category above the panel.
We estimate uncertainty with 2,000 bootstrap resamples of whole items,
retaining the dependence among tokens in each response.
Where available, we report $\pm$ one bootstrap
SD.

\subsection{Compute concentrates in a sparse tail}
\label{sec:landscape}

\begin{figure}[t]
\centering
\includegraphics[width=\textwidth]{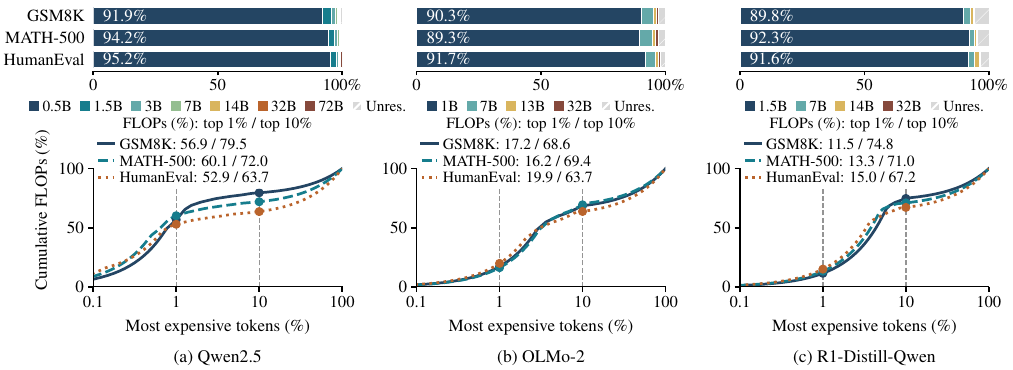}
\caption{Token reproduction and estimated FLOP concentration across
Qwen2.5, OLMo-2, and DeepSeek-R1-Distill-Qwen. \textit{Top}: distributions of the least-cost matching model for
GSM8K, MATH-500, and HumanEval.
Labels on the bar give the smallest-model share. Hatched gray segments denote
unresolved tokens (Unres.), which might require larger models to reproduce.
\textit{Bottom}: cumulative estimated FLOPs over all reference tokens, ordered
from most to least expensive, with a log horizontal axis.
Resolved tokens retain their measured architecture-derived FLOP costs;
each unresolved token is assigned twice the prediction FLOPs of the
largest model in its panel. Vertical dashed
lines and markers identify the top 1\% and 10\% of all tokens.}
\label{fig:concentration}
\end{figure}

\textbf{Most tokens match the smallest model and cost concentrates in the tail.}
Across all three measurement panels, most reference tokens are assigned
to the smallest available model (Shown in Figure~\ref{fig:concentration}).
On the same Qwen-generated references, the 0.5B Qwen model reproduces
91.86--95.24\% of tokens across GSM8K, MATH-500, and HumanEval.
The 1B OLMo and 1.5B R1-Distill-Qwen models reproduce 89.34--91.74\% and
89.80--92.27\%, respectively. The predominance of inexpensive predictions persists across all three model families.

The most expensive 10\% of tokens account for
63.66--79.54\% of estimated FLOPs in Qwen, 63.75--69.37\% in OLMo,
and 67.19--74.76\% in R1-Distill-Qwen. The top 1\% account for
52.94--60.14\%, 16.25--19.87\%, and 11.47--15.02\%, respectively.
The estimated compute is therefore concentrated in a small fraction
of positions across all three panels, with Qwen exhibiting a sharper
concentration within the most expensive percentile.
We next examine whether the panels agree on the ordering of expensive positions, with each panel supplying its own cost scale.

\textbf{Different families share a reproduction ordering.}
\begin{table}[t]
\centering
\caption{Model families share a reproduction ordering.
Partial Spearman correlations of aligned word-level reproduction indices,
pooled across GSM8K, MATH-500, and HumanEval, controlling for generator
surprisal, probe surprisal, and position. Columns distinguish reference
generators. Values are estimates $\pm$ one item-bootstrap SD.
A 0 indicates no rank association.}
\label{tab:agreement-summary}
\small
\setlength{\tabcolsep}{7pt}
\begin{tabular}{@{}lcc@{}}
\toprule
Model families & Qwen references & Llama references \\
\midrule
Qwen2.5 / OLMo-2 & $0.322 \pm 0.003$ & $0.417 \pm 0.003$ \\
Qwen2.5 / Falcon3 & $0.380 \pm 0.003$ & $0.475 \pm 0.003$ \\
OLMo-2 / Falcon3 & $0.398 \pm 0.003$ & $0.447 \pm 0.003$ \\
\bottomrule
\end{tabular}
\end{table}
\label{sec:agreement}
Different model families exhibit a shared ordering of reproduction difficulty. After controlling for generator surprisal, probe surprisal, and position, the partial rank correlation between each pair of families, pooled over the three core datasets, ranges from 0.322 to 0.398 on Qwen-generated references and from 0.417 to 0.475 on Llama-generated references (Table~\ref{tab:agreement-summary}). Positions that require larger models in one family tend to require them in another, and the pattern extends to text generated by another family. The association is positive on every benchmark and moves by less than 0.05 under additional controls, alternative success criteria, a second verified response per problem, memorization screening, and GSM-Symbolic (Appendix~\ref{app:robustness}, Tables~\ref{tab:dataset-correlations} and~\ref{tab:robustness}). It is also stronger than greedy or soft speculative-rejection signals on the same text (0.32/0.42 versus 0.15--0.38; Appendix~\ref{app:controls}).

\subsection{Cheap reproduction does not ensure reliable continuation}
\label{sec:handover}

The cost map measures token reproduction conditioned on the reference
prefix. To assess whether low reproduction costs translate into reliable
continuation, we allow smaller models to continue generating the remaining response and report accuracy drops relative to Qwen-72B.

\begin{figure}[t]
\centering
\includegraphics[width=\textwidth]{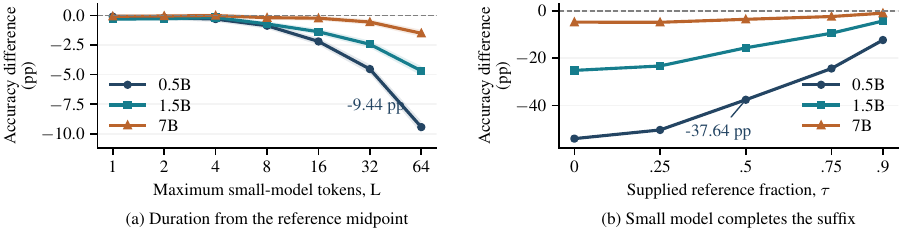}
\caption{Accuracy drops with the smaller model continuation.
Differences from a 72B continuation of the same verified prefix. \textit{(a)}: a small model generates up to $L$ tokens,
then 72B resumes. Shading shows $\pm$ one item-bootstrap SD.
\textit{(b)}: full-suffix generation from reference character fraction $\tau$.}
\label{fig:handovers}
\end{figure}

\textbf{Effect of continuation duration.}
We vary the duration of small-model generation before Qwen-72B resumes.  Both models condition
on the generated continuation without receiving additional reference
tokens. The baseline is an uninterrupted Qwen-72B continuation from the
same prefix.
Longer handovers produce larger accuracy losses, with the magnitude
depending on model capacity (Figure~\ref{fig:handovers}, left).
For the 0.5B model, the accuracy difference relative to Qwen-72B is
$-0.09\pm0.19$ percentage points at one token and
decreases to $-9.44\pm0.50$ at 64 tokens.
Larger models mitigate this degradation: at 64 tokens, the losses
are $4.68\pm0.39$ points for 1.5B and $1.50\pm0.30$ points for 7B.
These results motivate a joint choice of model capacity and delegation length: realizing the savings suggested by the sufficient-compute map requires accounting for how delegated
predictions affect the rest of the trajectory. Appendix~\ref{app:handover-details} specifies the detail.

\textbf{Effect of model capacity and reference-prefix length.}
We evaluate full-suffix continuation from reference fractions
$\tau\in\{0,0.25,0.5,0.75,0.9\}$, allowing the selected model to generate
the entire remaining response. Relative to Qwen-72B, the 0.5B model
loses 54.08 accuracy points when generating from the prompt
($\tau=0$) and 37.64 points when starting from the reference midpoint.



\subsection{Sufficient-compute maps reveal headroom for efficient inference}
\begin{figure}[t]
\centering
\includegraphics[width=\textwidth]{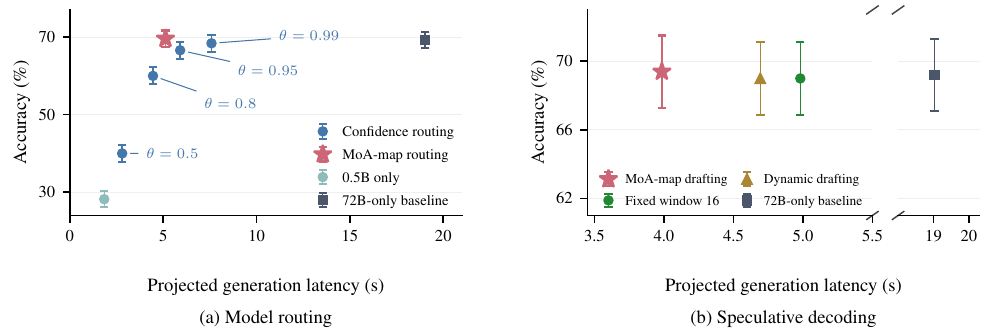}
\caption{Accuracy versus latency on MATH-500. (a) Confidence routing selects 0.5B when its maximum next-token
probability is at least $\theta$. MoA-map routing uses the sufficient-compute map.
(b) Draft lengths follow a fixed window, model confidence, or consecutive
sufficient-compute matches. Gray squares show the 72B-only baseline. Vertical bars show one item-bootstrap standard deviation. 
}
\label{fig:allocation}
\end{figure}

We evaluate whether the sufficient-compute map can guide model routing and
speculative decoding.
Figure~\ref{fig:allocation} compares measured answer accuracy with projected generation
latency on MATH-500. 
The map is supplied as an oracle: a greedy 72B response provides the reference for each problem, the controller receives the positions at which 0.5B reproduces it. Appendix~\ref{app:allocation-live} specifies the design. 

\textbf{Model routing.}
Our baseline, token-level cascade~\citep{narasimhan2025faster}, uses the
0.5B model when its maximum next-token probability reaches a threshold
and defers to 72B otherwise.
MoA-map routing replaces this confidence gate with measured agreement:
it selects 0.5B at positions where the map indicates that it reproduces
the token, and 72B elsewhere.
MoA-map routing achieves 69.60\% accuracy, compared with 68.40\% for the best-performing confidence setting ($\theta=0.99$), yielding the best accuracy–latency tradeoff (shown in Figure~\ref{fig:allocation}a).
The comparison identifies an opportunity beyond confidence-based routing:
token agreement can guide delegation to a smaller model while
retaining high answer accuracy. 

\textbf{Speculative decoding.}
The map can also determine how many tokens the small model should draft
before verification in speculative decoding~\citep{leviathan2023speculative}.
MoA-map drafting follows consecutive positions marked as reproducible
by 0.5B and sends the resulting block to 72B for verification.
It achieves 69.40\% accuracy, compared with 69.00\% for both a fixed
16-token window and confidence-adaptive drafting~\citep{mamou2024dynamicblog}.
Meanwhile, the projected latencies are 3.99, 4.98, and 4.69 seconds per response, reducing latency by up to 19.9\% (shown in Figure~\ref{fig:allocation}b).
MoA-map drafting generates 32.6\% fewer draft tokens than the fixed-window
baseline: anticipating disagreement reduces proposals that the target would
discard. This reduction in draft work explains its lower latency.
Together, the two interventions use sufficient-compute map to identify where
small-model predictions suffice and how long drafting should continue
before verification.

\section{Related work}
\label{sec:related}


\textbf{Efficient serving.}
Efficient LLM serving combines systems optimizations with adaptive computation.
Orca combines iteration-level scheduling with selective batching
\citep{yu2022orca}, while vLLM reduces KV-cache fragmentation and enables
memory sharing through PagedAttention \citep{kwon2023vllm}.
Speculative decoding preserves a target distribution by verifying draft
proposals \citep{leviathan2023speculative,chen2023accelerating}. Its acceptance
rate depends on alignment between distributions; a rejected proposal can
still be a valid continuation \citep{bachmann2025judge}. 
FrugalGPT and RouteLLM allocate requests across models to balance cost and
quality \citep{chen2023frugalgpt,ong2025routellm}, while token-level cascades
make small/large-model decisions within a response
\citep{narasimhan2025faster}.
At the architectural level, MoE increases parameter capacity through sparse
expert activation \citep{shazeer2017moe,fedus2022switch}.
Expert placement and load balancing
help translate this sparsity into serving efficiency
\citep{du2024sida,ma2025moegps}.
Our work complements these approaches by measuring which complete model can
reproduce each token given its reference prefix, then testing whether this
information improves model selection and speculative draft length.

\textbf{Token-level control.}
CITER and R2R learn small/large-model routing
\citep{zheng2025citer,fu2025r2r}, while COLLAB selects agents using
long-term utility \citep{chakraborty2025collab}.
Our instrument measures local reference cost before testing its usefulness
for these sequential decisions.
Duo-LLM studies oracle module allocation, FS-GEN studies small/large-model
collaboration, and FlexiDepth learns layer skipping
\citep{alizadeh2024duollm,zhang2024fsgen,luo2025flexidepth}.
CALM and Mixture-of-Depths also allocate computation within a model
\citep{schuster2022calm,raposo2024mod}.
Appendix~\ref{app:flexi} quantifies agreement between our reproduction indices and FlexiDepth's learned allocations; Appendix~\ref{app:related-full} relates our measurement design to these methods.

\textbf{Shared difficulty and trajectory interventions.}
Earlier work finds shared learning order across classifiers
\citep{hacohen2020agree} and studies token learning trajectories across model
scales \citep{xia2023trajectories,chang2024learningcurves}. Our cross-family
test asks whether a related ordering persists within verified generated text.
Its interpretation depends on the operational criterion \citep{jacobs2021measurement}.
Critical-token studies locate influential positions through contrasting
continuations \citep{lin2025critical}. 
Scheduled sampling and work on exposure bias
distinguish supplied-prefix prediction from free generation
\citep{bengio2015scheduled,he2021exposure,bachmann2024pitfalls}.
Model handovers also appear in Reasoning Relay, LLM Shepherding, and
RelayGen \citep{lu2025relay,dong2026shepherding,song2026relaygen}.
Our duration sweep relates these continuation interventions to a separately
measured reproduction cost.

\section{Discussion and limitations}
\label{sec:discussion}

\textbf{Discussion.}
Our study uses sufficient compute as an analytical reference for identifying efficiency headroom in existing serving methods. We show that better allocation information can improve the accuracy–efficiency trade-off.
However, constructing the map requires evaluating every candidate model along known reference sequences. Developing practical controllers that recover these benefits without exhaustive evaluation or
advance reference access remains an open problem.
The shared ordering of reproduction demands across model families raises a further question: are measured maps transferable? Cross-family agreement motivates this possibility but does not establish transferability. 

In our experiments, longer words show a weak positive association with reproduction demand, whereas digit-containing and operator/punctuation spans show weak negative associations. Regressions on digit and operator indicators, word length, frequency, and position explain only 2.5--3.2\% of the variance in ranked reproduction demand (Appendix~\ref{app:controls}). Thus, simple surface categories provide limited guidance about which tokens require larger models and identifying the contextual and semantic properties of the expensive tail could be an interesting direction.




\textbf{Limitations.}
Exact reproduction excludes alternative valid wordings and reasoning paths,
particularly in open-ended natural language. Meanwhile,
exact agreement may be more informative for structured outputs whose tokens
have fixed meanings, such as tokenized visual predictions.
The results are conditional on the evaluated model panels and benchmarks.                         
Most reproduction analyses use checker-passing references, limiting coverage                      
of unsuccessful trajectories and difficult problems.

\section{Conclusion}
\label{sec:conclusion}

We introduced a token-level MoA framework for measuring sufficient compute. Most reference tokens can be reproduced by the smallest model, while a sparse expensive tail accounts for disproportionate assigned computation. Shared reproduction orderings extend across
the tested families, but handover experiments show that local reproduction alone does not ensure reliable continuation. On MATH-500, oracle-guided routing and drafting improve projected efficiency at comparable accuracy, revealing remaining allocation headroom.
Exploiting this opportunity without advance reference access or exhaustive model evaluation remains a central challenge for practical inference.


\clearpage

\bibliographystyle{abbrvnat}
\bibliography{refs}

@inproceedings{leviathan2023speculative,
  title={Fast inference from transformers via speculative decoding},
  author={Leviathan, Yaniv and Kalman, Matan and Matias, Yossi},
  booktitle={International conference on machine learning},
  pages={19274--19286},
  year={2023},
  organization={PMLR}
}

@article{chen2023accelerating,
  title={Accelerating large language model decoding with speculative sampling},
  author={Chen, Charlie and Borgeaud, Sebastian and Irving, Geoffrey and Lespiau, Jean-Baptiste and Sifre, Laurent and Jumper, John},
  journal={arXiv preprint arXiv:2302.01318},
  year={2023}
}

@inproceedings{bachmann2025judge,
  title={Judge decoding: Faster speculative sampling requires going beyond model alignment},
  author={Bachmann, Gregor and Anagnostidis, Sotiris and Pumarola, Albert and Georgopoulos, Markos and Sanakoyeu, Artsiom and Du, Yuming and Sch{\"o}nfeld, Edgar and Thabet, Ali and Kohler, Jonas},
  booktitle={International Conference on Learning Representations},
  volume={2025},
  pages={40894--40914},
  year={2025}
}

@article{schuster2022calm,
  title={Confident adaptive language modeling},
  author={Schuster, Tal and Fisch, Adam and Gupta, Jai and Dehghani, Mostafa and Bahri, Dara and Tran, Vinh and Tay, Yi and Metzler, Donald},
  journal={Advances in Neural Information Processing Systems},
  volume={35},
  pages={17456--17472},
  year={2022}
}

@article{raposo2024mod,
  title={Mixture-of-depths: Dynamically allocating compute in transformer-based language models},
  author={Raposo, David and Ritter, Sam and Richards, Blake and Lillicrap, Timothy and Humphreys, Peter Conway and Santoro, Adam},
  journal={arXiv preprint arXiv:2404.02258},
  year={2024}
}

@article{luo2025flexidepth,
  title={Adaptive layer-skipping in pre-trained llms},
  author={Luo, Xuan and Wang, Weizhi and Yan, Xifeng},
  journal={arXiv preprint arXiv:2503.23798},
  year={2025}
}

@inproceedings{ong2025routellm,
  title        = {{RouteLLM}: Learning to Route {LLMs} with Preference Data},
  author       = {Ong, Isaac and Almahairi, Amjad and Wu, Vincent and others},
  booktitle    = {International Conference on Learning Representations (ICLR)},
  year         = {2025},
  eprint       = {2406.18665},
  archivePrefix= {arXiv},
  primaryClass = {cs.LG}
}

@misc{chen2023frugalgpt,
  title        = {{FrugalGPT}: How to Use Large Language Models While Reducing Cost and
                  Improving Performance},
  author       = {Chen, Lingjiao and Zaharia, Matei and Zou, James},
  year         = {2023},
  eprint       = {2305.05176},
  archivePrefix= {arXiv},
  primaryClass = {cs.LG}
}

@inproceedings{alizadeh2024duollm,
  title={Duo-LLM: A Framework for Studying Adaptive Computation in Large Language Models},
  author={Alizadeh-Vahid, Keivan and Mirzadeh, Seyed Iman and Shahrkokhi, Hooman and Belenko, Dmitry and Sun, Frank and Cho, Minsik and Sekhavat, Mohammad Hossein and Nabi, Moin and Farajtabar, Mehrdad},
  booktitle={NeurIPS Efficient Natural Language and Speech Processing Workshop},
  pages={443--455},
  year={2024},
  organization={PMLR}
}

@article{zhang2024fsgen,
  title={Fast and slow generating: An empirical study on large and small language models collaborative decoding},
  author={Zhang, Kaiyan and Wang, Jianyu and Ding, Ning and Qi, Biqing and Hua, Ermo and Lv, Xingtai and Zhou, Bowen},
  journal={arXiv preprint arXiv:2406.12295},
  year={2024}
}

@inproceedings{hacohen2020agree,
  title={Let’s agree to agree: Neural networks share classification order on real datasets},
  author={Hacohen, Guy and Choshen, Leshem and Weinshall, Daphna},
  booktitle={International Conference on Machine Learning},
  pages={3950--3960},
  year={2020},
  organization={PMLR}
}

@article{chang2024learningcurves,
  title={Characterizing learning curves during language model pre-training: Learning, forgetting, and stability},
  author={Chang, Tyler A and Tu, Zhuowen and Bergen, Benjamin K},
  journal={Transactions of the Association for Computational Linguistics},
  volume={12},
  pages={1346--1362},
  year={2024},
  publisher={MIT Press 255 Main Street, 9th Floor, Cambridge, Massachusetts 02142, USA~…}
}

@inproceedings{xia2023trajectories,
  title={Training trajectories of language models across scales},
  author={Xia, Mengzhou and Artetxe, Mikel and Zhou, Chunting and Lin, Xi Victoria and Pasunuru, Ramakanth and Chen, Danqi and Zettlemoyer, Luke and Stoyanov, Veselin},
  booktitle={Proceedings of the 61st Annual Meeting of the Association for Computational Linguistics (Volume 1: Long Papers)},
  pages={13711--13738},
  year={2023}
}

@inproceedings{
lin2025critical,
title={Critical Tokens Matter: Token-Level Contrastive Estimation Enhances {LLM}{\textquoteright}s Reasoning Capability},
author={Zicheng Lin and Tian Liang and Jiahao Xu and Qiuzhi Liu and Xing Wang and Ruilin Luo and Chufan Shi and Siheng Li and Yujiu Yang and Zhaopeng Tu},
booktitle={Forty-second International Conference on Machine Learning},
year={2025},
url={https://openreview.net/forum?id=fnz1g18EdI}
}

@inproceedings{jacobs2021measurement,
  title={Measurement and fairness},
  author={Jacobs, Abigail Z and Wallach, Hanna},
  booktitle={Proceedings of the 2021 ACM conference on fairness, accountability, and transparency},
  pages={375--385},
  year={2021}
}

@inproceedings{bengio2015scheduled,
 author = {Bengio, Samy and Vinyals, Oriol and Jaitly, Navdeep and Shazeer, Noam},
 booktitle = {Advances in Neural Information Processing Systems},
 editor = {C. Cortes and N. Lawrence and D. Lee and M. Sugiyama and R. Garnett},
 pages = {},
 publisher = {Curran Associates, Inc.},
 title = {Scheduled Sampling for Sequence Prediction with Recurrent Neural Networks},
 url = {https://proceedings.neurips.cc/paper_files/paper/2015/file/e995f98d56967d946471af29d7bf99f1-Paper.pdf},
 volume = {28},
 year = {2015}
}

@inproceedings{
bachmann2024pitfalls,
title={The Pitfalls of Next-Token Prediction},
author={Gregor Bachmann and Vaishnavh Nagarajan},
booktitle={Forty-first International Conference on Machine Learning},
year={2024},
url={https://openreview.net/forum?id=76zq8Wkl6Z}
}

@inproceedings{he2021exposure,
    title = "Exposure Bias versus Self-Recovery: Are Distortions Really Incremental for Autoregressive Text Generation?",
    author = "He, Tianxing  and
      Zhang, Jingzhao  and
      Zhou, Zhiming  and
      Glass, James",
    editor = "Moens, Marie-Francine  and
      Huang, Xuanjing  and
      Specia, Lucia  and
      Yih, Scott Wen-tau",
    booktitle = "Proceedings of the 2021 Conference on Empirical Methods in Natural Language Processing",
    month = nov,
    year = "2021",
    address = "Online and Punta Cana, Dominican Republic",
    publisher = "Association for Computational Linguistics",
    url = "https://aclanthology.org/2021.emnlp-main.415/",
    doi = "10.18653/v1/2021.emnlp-main.415",
    pages = "5087--5102"
}

@misc{qwen2024qwen25,
      title={Qwen2.5 Technical Report}, 
      author={Qwen and : and An Yang and Baosong Yang and Beichen Zhang and Binyuan Hui and Bo Zheng and Bowen Yu and Chengyuan Li and Dayiheng Liu and Fei Huang and Haoran Wei and Huan Lin and Jian Yang and Jianhong Tu and Jianwei Zhang and Jianxin Yang and Jiaxi Yang and Jingren Zhou and Junyang Lin and Kai Dang and Keming Lu and Keqin Bao and Kexin Yang and Le Yu and Mei Li and Mingfeng Xue and Pei Zhang and Qin Zhu and Rui Men and Runji Lin and Tianhao Li and Tianyi Tang and Tingyu Xia and Xingzhang Ren and Xuancheng Ren and Yang Fan and Yang Su and Yichang Zhang and Yu Wan and Yuqiong Liu and Zeyu Cui and Zhenru Zhang and Zihan Qiu},
      year={2025},
      eprint={2412.15115},
      archivePrefix={arXiv},
      primaryClass={cs.CL},
      url={https://arxiv.org/abs/2412.15115}, 
}

@misc{olmo2025olmo2,
      title={2 OLMo 2 Furious}, 
      author={Team OLMo and Pete Walsh and Luca Soldaini and Dirk Groeneveld and Kyle Lo and Shane Arora and Akshita Bhagia and Yuling Gu and Shengyi Huang and Matt Jordan and Nathan Lambert and Dustin Schwenk and Oyvind Tafjord and Taira Anderson and David Atkinson and Faeze Brahman and Christopher Clark and Pradeep Dasigi and Nouha Dziri and Allyson Ettinger and Michal Guerquin and David Heineman and Hamish Ivison and Pang Wei Koh and Jiacheng Liu and Saumya Malik and William Merrill and Lester James V. Miranda and Jacob Morrison and Tyler Murray and Crystal Nam and Jake Poznanski and Valentina Pyatkin and Aman Rangapur and Michael Schmitz and Sam Skjonsberg and David Wadden and Christopher Wilhelm and Michael Wilson and Luke Zettlemoyer and Ali Farhadi and Noah A. Smith and Hannaneh Hajishirzi},
      year={2025},
      eprint={2501.00656},
      archivePrefix={arXiv},
      primaryClass={cs.CL},
      url={https://arxiv.org/abs/2501.00656}, 
}

@article{grattafiori2024llama3,
  title={The llama 3 herd of models},
  author={Grattafiori, Aaron and Dubey, Abhimanyu and Jauhri, Abhinav and Pandey, Abhinav and Kadian, Abhishek and Al-Dahle, Ahmad and Letman, Aiesha and Mathur, Akhil and Schelten, Alan and Vaughan, Alex and others},
  journal={arXiv preprint arXiv:2407.21783},
  year={2024}
}

@article{cobbe2021gsm8k,
  title={Training verifiers to solve math word problems},
  author={Cobbe, Karl and Kosaraju, Vineet and Bavarian, Mohammad and Chen, Mark and Jun, Heewoo and Kaiser, Lukasz and Plappert, Matthias and Tworek, Jerry and Hilton, Jacob and Nakano, Reiichiro and others},
  journal={arXiv preprint arXiv:2110.14168},
  year={2021}
}

@inproceedings{lightman2023verify,
  title={Let's verify step by step},
  author={Lightman, Hunter and Kosaraju, Vineet and Burda, Yuri and Edwards, Harrison and Baker, Bowen and Lee, Teddy and Leike, Jan and Schulman, John and Sutskever, Ilya and Cobbe, Karl},
  booktitle={International Conference on Learning Representations},
  volume={2024},
  pages={39578--39601},
  year={2024}
}

@article{chen2021humaneval,
  title={Evaluating large language models trained on code},
  author={Chen, Mark and Tworek, Jerry and Jun, Heewoo and Yuan, Qiming and Pinto, Henrique Ponde De Oliveira and Kaplan, Jared and Edwards, Harri and Burda, Yuri and Joseph, Nicholas and Brockman, Greg and others},
  journal={arXiv preprint arXiv:2107.03374},
  year={2021}
}

@inproceedings{jain2024livecodebench,
  title={Livecodebench: Holistic and contamination free evaluation of large language models for code},
  author={Jain, Naman and Gu, Alex and Li, Wen-Ding and Yan, Fanjia and Zhang, Tianjun and Wang, Sida and Solar-Lezama, Armando and Sen, Koushik and Stoica, Ion},
  booktitle={International Conference on Learning Representations},
  volume={2025},
  pages={58791--58831},
  year={2025}
}

@article{kwiatkowski2019nq,
    title = "Natural Questions: A Benchmark for Question Answering Research",
    author = "Kwiatkowski, Tom  and
      Palomaki, Jennimaria  and
      Redfield, Olivia  and
      Collins, Michael  and
      Parikh, Ankur  and
      Alberti, Chris  and
      Epstein, Danielle  and
      Polosukhin, Illia  and
      Devlin, Jacob  and
      Lee, Kenton  and
      Toutanova, Kristina  and
      Jones, Llion  and
      Kelcey, Matthew  and
      Chang, Ming-Wei  and
      Dai, Andrew M.  and
      Uszkoreit, Jakob  and
      Le, Quoc  and
      Petrov, Slav",
    editor = "Lee, Lillian  and
      Johnson, Mark  and
      Roark, Brian  and
      Nenkova, Ani",
    journal = "Transactions of the Association for Computational Linguistics",
    volume = "7",
    year = "2019",
    address = "Cambridge, MA",
    publisher = "MIT Press",
    url = "https://aclanthology.org/Q19-1026/",
    doi = "10.1162/tacl_a_00276",
    pages = "452--466"
}

@inproceedings{kwon2023vllm,
  title={Efficient memory management for large language model serving with pagedattention},
  author={Kwon, Woosuk and Li, Zhuohan and Zhuang, Siyuan and Sheng, Ying and Zheng, Lianmin and Yu, Cody Hao and Gonzalez, Joseph and Zhang, Hao and Stoica, Ion},
  booktitle={Proceedings of the 29th symposium on operating systems principles},
  pages={611--626},
  year={2023}
}

@misc{kydlicek2025mathverify,
  author = {Hynek Kydlíček},
  title = {math-verify: Mathematical Expression Verification},
  url = {https://github.com/huggingface/Math-Verify},
  year = {2025}
}

@inproceedings{
lu2025relay,
title={Evaluating Stability and Interchangeability of Large Language Models in Mathematical Reasoning},
author={Leo Lu and Jonathan Zhang and Sean Eugene Chua and Spencer Kim and Kevin Zhu and Sean O'Brien and Vasu Sharma},
booktitle={Socially Responsible and Trustworthy Foundation Models at NeurIPS 2025},
year={2025},
url={https://openreview.net/forum?id=QqXPAuKi76}
}

@article{dong2026shepherding,
  title={Pay for Hints, Not Answers: LLM Shepherding for Cost-Efficient Inference},
  author={Dong, Ziming and Sharma, Hardik and O'Toole, Evan and Champati, Jaya Prakash and Wu, Kui},
  journal={arXiv preprint arXiv:2601.22132},
  year={2026}
}

@article{song2026relaygen,
  title={RelayGen: Intra-Generation Model Switching for Efficient Reasoning},
  author={Song, Jiwon and Kim, Yoongon and Kim, Jae-Joon},
  journal={arXiv preprint arXiv:2602.06454},
  year={2026}
}

@inproceedings{mirzadeh2024gsmsymbolic,
  title={Gsm-symbolic: Understanding the limitations of mathematical reasoning in large language models},
  author={Mirzadeh, Iman and Alizadeh-Vahid, Keivan and Shahrokhi, Hooman and Tuzel, Oncel and Bengio, Samy and Farajtabar, Mehrdad},
  booktitle={International Conference on Learning Representations},
  volume={2025},
  pages={94743--94765},
  year={2025}
}

@inproceedings{shen2024colllm,
  title={Learning to decode collaboratively with multiple language models},
  author={Shen, Zejiang and Lang, Hunter and Wang, Bailin and Kim, Yoon and Sontag, David},
  booktitle={Proceedings of the 62nd Annual Meeting of the Association for Computational Linguistics (Volume 1: Long Papers)},
  pages={12974--12990},
  year={2024}
}

@article{zheng2025citer,
  title={Citer: Collaborative inference for efficient large language model decoding with token-level routing},
  author={Zheng, Wenhao and Chen, Yixiao and Zhang, Weitong and Kundu, Souvik and Li, Yun and Liu, Zhengzhong and Xing, Eric P and Wang, Hongyi and Yao, Huaxiu},
  journal={arXiv preprint arXiv:2502.01976},
  year={2025}
}

@inproceedings{
fu2025r2r,
title={R2R: Efficiently Navigating Divergent Reasoning Paths with Small-Large Model Token Routing},
author={Tianyu Fu and Yi Ge and Yichen You and Enshu Liu and Zhihang Yuan and Guohao Dai and Shengen Yan and Huazhong Yang and Yu Wang},
booktitle={The Thirty-ninth Annual Conference on Neural Information Processing Systems},
year={2025},
url={https://openreview.net/forum?id=DpeJYRFRQY}
}

@inproceedings{
wang2024moa,
title={Mixture-of-Agents Enhances Large Language Model Capabilities},
author={Junlin Wang and Jue WANG and Ben Athiwaratkun and Ce Zhang and James Zou},
booktitle={The Thirteenth International Conference on Learning Representations},
year={2025},
url={https://openreview.net/forum?id=h0ZfDIrj7T}
}

@inproceedings{
chakraborty2025collab,
title={Collab: Controlled Decoding using Mixture of Agents for {LLM} Alignment},
author={Souradip Chakraborty and Sujay Bhatt and Udari Madhushani Sehwag and Soumya Suvra Ghosal and Jiahao Qiu and Mengdi Wang and Dinesh Manocha and Furong Huang and Alec Koppel and Sumitra Ganesh},
booktitle={The Thirteenth International Conference on Learning Representations},
year={2025},
url={https://openreview.net/forum?id=7ohlQUbTpp}
}

@article{clark2018arc,
  title={Think you have solved question answering? try arc, the ai2 reasoning challenge},
  author={Clark, Peter and Cowhey, Isaac and Etzioni, Oren and Khot, Tushar and Sabharwal, Ashish and Schoenick, Carissa and Tafjord, Oyvind},
  journal={arXiv preprint arXiv:1803.05457},
  year={2018}
}

@inproceedings{zellers2019hellaswag,
  title={Hellaswag: Can a machine really finish your sentence?},
  author={Zellers, Rowan and Holtzman, Ari and Bisk, Yonatan and Farhadi, Ali and Choi, Yejin},
  booktitle={Proceedings of the 57th annual meeting of the association for computational linguistics},
  pages={4791--4800},
  year={2019}
}

@article{liu2023evalplus,
  title={Is your code generated by chatgpt really correct? rigorous evaluation of large language models for code generation},
  author={Liu, Jiawei and Xia, Chunqiu Steven and Wang, Yuyao and Zhang, Lingming},
  journal={Advances in neural information processing systems},
  volume={36},
  pages={21558--21572},
  year={2023}
}

@inproceedings{suzgun2022bbh,
  title={Challenging big-bench tasks and whether chain-of-thought can solve them},
  author={Suzgun, Mirac and Scales, Nathan and Sch{\"a}rli, Nathanael and Gehrmann, Sebastian and Tay, Yi and Chung, Hyung Won and Chowdhery, Aakanksha and Le, Quoc and Chi, Ed H and Zhou, Denny and others},
  booktitle={Findings of the Association for Computational Linguistics: ACL 2023},
  pages={13003--13051},
  year={2023}
}

@article{hendrycks2021mmlu,
  title={Measuring massive multitask language understanding},
  author={Hendrycks, Dan and Burns, Collin and Basart, Steven and Zou, Andy and Mazeika, Mantas and Song, Dawn and Steinhardt, Jacob},
  journal={arXiv preprint arXiv:2009.03300},
  year={2020}
}

@article{wang2024mmlupro,
  title={Mmlu-pro: A more robust and challenging multi-task language understanding benchmark},
  author={Wang, Yubo and Ma, Xueguang and Zhang, Ge and Ni, Yuansheng and Chandra, Abhranil and Guo, Shiguang and Ren, Weiming and Arulraj, Aaran and He, Xuan and Jiang, Ziyan and others},
  journal={Advances in Neural Information Processing Systems},
  volume={37},
  pages={95266--95290},
  year={2024}
}

@article{fedus2022switch,
  title={Switch transformers: Scaling to trillion parameter models with simple and efficient sparsity},
  author={Fedus, William and Zoph, Barret and Shazeer, Noam},
  journal={Journal of Machine Learning Research},
  volume={23},
  number={120},
  pages={1--39},
  year={2022}
}

@inproceedings{
shazeer2017moe,
title={ Outrageously Large Neural Networks: The Sparsely-Gated Mixture-of-Experts Layer},
author={Noam Shazeer and *Azalia Mirhoseini and *Krzysztof Maziarz and Andy Davis and Quoc Le and Geoffrey Hinton and Jeff Dean},
booktitle={International Conference on Learning Representations},
year={2017},
url={https://openreview.net/forum?id=B1ckMDqlg}
}

@article{ranzato2016sequence,
  title={Sequence level training with recurrent neural networks},
  author={Ranzato, Marc'Aurelio and Chopra, Sumit and Auli, Michael and Zaremba, Wojciech},
  journal={arXiv preprint arXiv:1511.06732},
  year={2015}
}

@inproceedings{
bahdanau2017actor,
title={An Actor-Critic Algorithm for Sequence Prediction},
author={Dzmitry Bahdanau and Philemon Brakel and Kelvin Xu and Anirudh Goyal and Ryan Lowe and Joelle Pineau and Aaron Courville and Yoshua Bengio},
booktitle={International Conference on Learning Representations},
year={2017},
url={https://openreview.net/forum?id=SJDaqqveg}
}

@inproceedings{
mudgal2024controlled,
title={Controlled Decoding from Language Models},
author={Sidharth Mudgal and Jong Lee and Harish Ganapathy and YaGuang Li and Tao Wang and Yanping Huang and Zhifeng Chen and Heng-Tze Cheng and Michael Collins and Trevor Strohman and Jilin Chen and Alex Beutel and Ahmad Beirami},
booktitle={Forty-first International Conference on Machine Learning},
year={2024},
url={https://openreview.net/forum?id=bVIcZb7Qa0}
}

@article{chakraborty2024transfer,
  title={Transfer q-star: Principled decoding for llm alignment},
  author={Chakraborty, Souradip and Ghosal, Soumya Suvra and Yin, Ming and Manocha, Dinesh and Wang, Mengdi and Bedi, Amrit Singh and Huang, Furong},
  journal={Advances in Neural Information Processing Systems},
  volume={37},
  pages={101725--101761},
  year={2024}
}

@article{du2024sida,
  title={Sida: Sparsity-inspired data-aware serving for efficient and scalable large mixture-of-experts models},
  author={Du, Zhixu and Li, Shiyu and Wu, Yuhao and Jiang, Xiangyu and Sun, Jingwei and Zheng, Qilin and Wu, Yongkai and Li, Ang and Li, Hai H and Chen, Yiran},
  journal={Proceedings of Machine Learning and Systems},
  volume={6},
  pages={224--238},
  year={2024}
}

@inproceedings{
ma2025moegps,
title={MoE-{GPS}: Guidlines for Prediction Strategy with Expert Duplication in MoE Load Balancing},
author={Haiyue Ma and Zhixu Du and Yiran Chen},
booktitle={Machine Learning for Systems 2025},
year={2025},
url={https://openreview.net/forum?id=VCh5vBSxxb}
}

@article{jacobs1991mixtures,
  title={Adaptive mixtures of local experts},
  author={Jacobs, Robert A and Jordan, Michael I and Nowlan, Steven J and Hinton, Geoffrey E},
  journal={Neural computation},
  volume={3},
  number={1},
  pages={79--87},
  year={1991},
  publisher={MIT Press}
}

@article{brown2020fewshot,
  title={Language models are few-shot learners},
  author={Brown, Tom and Mann, Benjamin and Ryder, Nick and Subbiah, Melanie and Kaplan, Jared D and Dhariwal, Prafulla and Neelakantan, Arvind and Shyam, Pranav and Sastry, Girish and Askell, Amanda and others},
  journal={Advances in neural information processing systems},
  volume={33},
  pages={1877--1901},
  year={2020}
}

@article{openai2023gpt4,
  title={Gpt-4 technical report},
  author={Achiam, Josh and Adler, Steven and Agarwal, Sandhini and Ahmad, Lama and Akkaya, Ilge and Aleman, Florencia Leoni and Almeida, Diogo and Altenschmidt, Janko and Altman, Sam and Anadkat, Shyamal and others},
  journal={arXiv preprint arXiv:2303.08774},
  year={2023}
}

@article{pope2023inference,
  title={Efficiently scaling transformer inference},
  author={Pope, Reiner and Douglas, Sholto and Chowdhery, Aakanksha and Devlin, Jacob and Bradbury, James and Heek, Jonathan and Xiao, Kefan and Agrawal, Shivani and Dean, Jeff},
  journal={Proceedings of machine learning and systems},
  volume={5},
  pages={606--624},
  year={2023}
}

@techreport{chatterji2025usage,
  title={How people use chatgpt},
  author={Chatterji, Aaron and Cunningham, Thomas and Deming, David J and Hitzig, Zoe and Ong, Christopher and Shan, Carl Yan and Wadman, Kevin},
  year={2025},
  institution={National Bureau of Economic Research}
}

@misc{falcon2024falcon3,
    title = {The Falcon 3 Family of Open Models},
    url = {https://huggingface.co/blog/falcon3},
    author = {Falcon-LLM Team},
    month = {December},
    year = {2024}
}

@article{guo2025deepseek,
  title={Deepseek-r1: Incentivizing reasoning capability in llms via reinforcement learning},
  author={Guo, Daya and Yang, Dejian and Zhang, Haowei and Song, Junxiao and Wang, Peiyi and Zhu, Qihao and Xu, Runxin and Zhang, Ruoyu and Ma, Shirong and Bi, Xiao and others},
  journal={arXiv preprint arXiv:2501.12948},
  year={2025}
}

@inproceedings{yu2022orca,
  title={Orca: A distributed serving system for $\{$Transformer-Based$\}$ generative models},
  author={Yu, Gyeong-In and Jeong, Joo Seong and Kim, Geon-Woo and Kim, Soojeong and Chun, Byung-Gon},
  booktitle={16th USENIX symposium on operating systems design and implementation (OSDI 22)},
  pages={521--538},
  year={2022}
}

@inproceedings{narasimhan2025faster,
  title={Faster cascades via speculative decoding},
  author={Narasimhan, Harikrishna and Jitkrittum, Wittawat and Rawat, Ankit Singh and Kim, Seungyeon and Gupta, Neha and Menon, Aditya Krishna and Kumar, Sanjiv},
  booktitle={International Conference on Learning Representations},
  volume={2025},
  pages={44949--44987},
  year={2025}
}

@misc{mamou2024dynamicblog,
  title = {Faster Assisted Generation with Dynamic Speculation},
  author = {Mamou, Jonathan and Pereg, Oren and Gante, Joao and Tunstall, Lewis and Korat, Daniel and Timor, Nadav and Wasserblat, Moshe},
  year = {2024},
  month = oct,
  howpublished = {Hugging Face Blog},
  url = {https://huggingface.co/blog/dynamic_speculation_lookahead}
}

@misc{qwen25speed,
  author = {{Qwen Team}},
  year = {n.d.},
  title = {{Qwen2.5} Speed Benchmark},
  howpublished = {\url{https://qwen.readthedocs.io/en/v2.5/benchmark/speed_benchmark.html}},
  note = {Official BF16 vLLM 0.6.3 benchmarks on A100 80GB GPUs; accessed 2026-09-25}
}

\clearpage
\appendix
\setcounter{table}{0}
\renewcommand{\thetable}{\textbf{A\arabic{table}}}
\section{Datasets, models, and selected populations}
\label{app:data}

\textbf{Benchmark coverage.}
Table~\ref{tab:datasets} separates attempted prompts, each generator's verified
pool, and their intersection. The primary collection contains twelve datasets
and 50,019 prompts. GSM-Symbolic adds 5,000 prompts as a separate robustness
analysis. Each core dataset has two-generator intersection coverage of at
least 60\%. The expanded collection is
reported per dataset and is not added to the pooled core estimand.

\begin{table}[!t]
\centering\small
\caption{Benchmark coverage before and after verification. The first three rows define the core. Counts refer to items.}
\label{tab:datasets}
\begin{tabular}{lrrrr}
\toprule
Dataset & Attempted & Qwen verified & Llama verified & Intersection \\
\midrule
GSM8K & 1,319 & 1,281 & 1,288 & 1,275 \\
MATH-500 & 500 & 375 & 363 & 347 \\
HumanEval & 164 & 144 & 146 & 136 \\
LiveCodeBench subset & 219 & 108 & 105 & 84 \\
NQ-Open & 3,610 & 1,090 & 1,355 & 862 \\
MBPP+ & 378 & 310 & 294 & 282 \\
AIME 2024 & 30 & 6 & 9 & 5 \\
ARC-Challenge & 1,172 & 1,137 & 1,136 & 1,120 \\
BBH & 6,511 & 5,509 & 5,647 & 5,355 \\
MMLU-Pro & 12,032 & 8,700 & 8,803 & 7,853 \\
MMLU & 14,042 & 12,303 & 12,260 & 11,611 \\
HellaSwag & 10,042 & 9,019 & 8,883 & 8,306 \\
\bottomrule
\end{tabular}
\end{table}

GSM8K uses the test split, MATH-500 the released 500-problem evaluation set,
and HumanEval the 164 official function-completion problems.
LiveCodeBench uses 219 problems with usable stdin/stdout tests from the
400-problem mirror available to the pipeline. Both public and private stored
tests are included. This is a subset, without a post-training-cutoff guarantee
\citep{jain2024livecodebench}. NQ-Open uses 3,610 validation questions with
normalized exact-match answer aliases \citep{kwiatkowski2019nq}.

\begingroup\raggedright
The extensions use \texttt{evalplus/mbppplus} (378 test items;
\citealp{liu2023evalplus}),
\texttt{HuggingFaceH4/aime\_2024} (30 problems),
\texttt{allenai/ai2\_arc} (ARC-Challenge test; \citealp{clark2018arc}),
\texttt{lukaemon/bbh} (the loaded task test splits; \citealp{suzgun2022bbh}),
\texttt{TIGER-Lab/MMLU-Pro} (test; \citealp{wang2024mmlupro}),
\texttt{cais/mmlu} (all subjects, test; \citealp{hendrycks2021mmlu}),
and \texttt{Rowan/hellaswag} (validation; \citealp{zellers2019hellaswag}).
BBH contains varied answer formats,
not exclusively multiple-choice questions. MBPP+ prompts include a supplied
assertion specifying the required function call.
GSM-Symbolic uses the released main split \citep{mirzadeh2024gsmsymbolic};
its transformed problems provide a replication substrate, but their public
release does not establish absence from model training.\par
\endgroup

\textbf{Prompts and checkers.}
Mathematical prompts request step-by-step reasoning and a final answer;
code prompts request executable code; knowledge and reasoning prompts request
an answer in the format specified by the dataset adapter.
GSM8K extracts a final numeric answer and compares it with numeric tolerance
$10^{-6}$. MATH-500 uses symbolic answer equivalence through Math-Verify
\citep{kydlicek2025mathverify}. HumanEval executes extracted code with the
stored official tests and a timeout. LiveCodeBench runs the stored stdin/stdout
cases; MBPP+ runs the executable tests available in the loaded records.
The multiple-choice adapters parse the final option, BBH uses task-answer
matching, and NQ-Open normalizes case, articles, punctuation, and whitespace
before exact alias matching. These are outcome checkers: intermediate reasoning
is retained even when a valid final answer follows an incorrect step.

\textbf{Generation settings.}
Reference sampling uses $K=4$, temperature 0.7, top-$p$ 0.95, and seed
20260729. Output caps are 512 tokens for GSM8K, HumanEval, MBPP+, ARC-Challenge,
and MMLU; 1,024 for MATH-500 and LiveCodeBench; 64 for NQ-Open; 2,048 for AIME;
768 for BBH and MMLU-Pro; and 384 for HellaSwag. GSM-Symbolic uses 512.
The earliest passing sample is selected independently for each generator.
A selected reference can be non-modal under its own generator because it
was sampled. Generation and scoring use bfloat16 with vLLM
\citep{kwon2023vllm}; the project runtime records version 0.14.1.

\subsection{Model panel and auxiliary models}
\label{app:models}
Table~\ref{tab:models} summarizes model roles and sizes. Qwen2.5,
OLMo-2, and R1-Distill-Qwen supply the fifteen candidate models displayed
in Figure~\ref{fig:concentration}
\citep{qwen2024qwen25,olmo2025olmo2,guo2025deepseek}.
The cross-family agreement analysis in Table~\ref{tab:agreement-summary}
uses Qwen2.5, OLMo-2, and the additional Falcon3 panel
\citep{falcon2024falcon3}. These are distinct panel comparisons.

\begin{table}[t]
\centering\small
\caption{Model roles and sizes. Measurement panels supply candidate
next-token policies; the Falcon panel tests cross-family agreement.
Reference generators supply candidate responses for task verification;
the additional MoE generator tests sensitivity to the reference source.
The probability probe supplies surprisal controls. Sizes are nominal
billions of parameters; the final row distinguishes total and active
parameters. Qwen-72B serves both as a panel member and a generator.}
\label{tab:models}
\begin{tabular}{@{}lll@{}}
\toprule
Role & Model family & Sizes (B) \\
\midrule
Measurement panel & Qwen2.5-Instruct & 0.5, 1.5, 3, 7, 14, 32, 72 \\
Measurement panel & OLMo-2-Instruct & 1, 7, 13, 32 \\
Reasoning-distilled panel & R1-Distill-Qwen & 1.5, 7, 14, 32 \\
Cross-family replication & Falcon3-Instruct & 1, 3, 7, 10 \\
\midrule
Probability probe & Llama-3.1-Instruct & 8 \\
Reference generator & Qwen2.5-Instruct & 72 \\
Reference generator & Llama-3.3-Instruct & 70 \\
Additional generator & Qwen3-30B-A3B & 30 total, 3 active \\
\bottomrule
\end{tabular}
\end{table}

\begingroup\raggedright
Qwen model identifiers have the form \texttt{Qwen/Qwen2.5-\{size\}B-Instruct}.
The OLMo identifiers are \texttt{allenai/OLMo-2-0425-1B-Instruct},
\texttt{allenai/OLMo-2-1124-7B-Instruct},
\texttt{allenai/OLMo-2-1124-13B-Instruct}, and
\texttt{allenai/OLMo-2-0325-32B-Instruct}.
Falcon identifiers are \texttt{tiiuae/Falcon3-\{size\}B-Instruct}.
The second main generator is \texttt{meta-llama/Llama-3.3-70B-Instruct};
the probability probe is \texttt{meta-llama/Llama-3.1-8B-Instruct}.

The reasoning-distilled panel contains four DeepSeek-R1-Distill-Qwen models,
with 1.5, 7, 14, and 32B parameters. Their identifiers follow
\path{deepseek-ai/DeepSeek-R1-Distill-Qwen-<size>B}.
These models test sensitivity to reasoning distillation. A third generator, \path{Qwen/Qwen3-30B-A3B}, supplies
mixture-of-experts reference text with thinking mode disabled. Its nominal
30B total and 3B active parameter counts describe architecture, not a measured
change in per-token compute allocation. These extensions use the three core
datasets.\par
\endgroup

\section{Measurement and statistical details}
\label{app:methods}

\textbf{Teacher-forced scoring.}
For each model, the user prompt is rendered using its chat template and
assistant-generation header. The reference answer is tokenized separately
without special tokens and appended to the prompt tokens. vLLM returns
prompt log probabilities for the answer positions. The cache stores target
rank, target log probability, argmax identity, character offsets, and a
truncated distribution. Log probabilities are rounded to four decimal places.
Each model receives the reference prefix even when its own preceding prediction
disagrees with the reference. Every model is evaluated at every position;
the algorithm does not stop after a smaller model matches.
The primary pass indicator is target rank one. Alternative criteria use
rank at most five or target log probability at least $\log(0.5)$, compared
at the stored float32 precision.

The entropy control sums $-p\log p$ over the returned top-twenty entries and
the target entry when additionally returned. It is an unnormalized truncated
entropy proxy.

\textbf{Word alignment.}
The alignment routine identifies non-whitespace runs and removes leading and
trailing ASCII punctuation, retaining all-punctuation spans. A native token
is assigned to the word with maximal character overlap; ties go to the
first word, and zero-overlap tokens are dropped. Only common mapped words
are joined across families. At word $w$, the ordinal statistic is
\begin{equation}
 D_F(w)=\max_{t:\,t\mapsto w} R_F(t),\qquad
 R_F(t)=\min\{k:a_{F,k,t}=1\}.
\end{equation}
An unresolved token gives the word the unresolved index above the panel.
The maximum summarizes constituent token indices; because predictions need
not be monotone in size, it does not certify that a single model reproduces
all constituent tokens. Surprisal uses sums over assigned tokens.

\textbf{Partial rank correlation.}
Let $r(\cdot)$ assign average ranks to ties and let
\[
 Z=[\mathbf{1},r(S_{\rm gen}),r(S_{\rm probe}),r(\log(1+\mathrm{position}))].
\]
Writing $P_Z$ for least-squares projection onto the columns of $Z$, we compute
\begin{equation}
 \rho_{\rm partial}=
 \operatorname{corr}\big((I-P_Z)r(D_Q),(I-P_Z)r(D_O)\big).
\end{equation}
Tables report the estimate $\pm$ the SD of 2,000 item-bootstrap estimates.
This bootstrap SD estimates the standard error under item sampling; models
and generation settings are held fixed. Main-text estimates and figure bars
or bands use the same bootstrap-SD convention where these SDs are available.
Point-only results omit uncertainty when no SD is available.
The bootstrap samples item identifiers with replacement and retains all
word rows belonging to each sampled item, including multiplicities.
Ranks and regressions are recomputed for each resample. Core Qwen text
contributes 319,336 words from 1,800 items; Llama text contributes 302,959
words from 1,797 items. The intersection sensitivity retains 307,850 and
293,549 words, respectively, from 1,758 items, giving point estimates of
0.319 and 0.415.


\subsection{Architecture-derived FLOP accounting}
\label{app:flops}

We count matrix-multiplication FLOPs from the released model configurations,
with one multiply-add counted as two operations. The count includes attention
projections, the two attention matrix products, gated feed-forward layers,
MoE router projections, and the vocabulary projection. Embedding lookup,
normalization, activations, bias additions, rotary embeddings, softmax, and
expert dispatch/combine are outside this arithmetic convention. Counts describe
logical causal operations; they do not include kernel padding, communication,
or memory movement.

Let $h$ be hidden width, $q$ the total query-head width, $r$ the total
key/value-head width, $L$ the number of layers, and $V$ vocabulary size.
For layer $\ell$, let $I_\ell^{\rm act}$ be the sum of the intermediate widths
of the feed-forward networks activated for one token. A dense gated MLP has
$I_\ell^{\rm act}=I_\ell$; a sparse layer with $a_\ell$ selected equal-sized
experts has $I_\ell^{\rm act}=a_\ell I_\ell$. Let $E_\ell$ denote the number
of experts scored by a router, and set $E_\ell=0$ in a dense layer. Define
\begin{align}
 A &= \sum_{\ell=1}^L
       \left[4h(q+r)+6hI_\ell^{\rm act}+2hE_\ell\right],
       & B&=4Lq, & H&=2hV.
 \label{eq:architecture-terms}
\end{align}
The first term counts query, key, value, and output projections; the second
counts the gate, up, and down feed-forward matrices. The router multiplies
an $h$-dimensional token by an $h\times E_\ell$ matrix. The two attention
products together require $B$ FLOPs per query/key position pair across layers.
For a prefix of $p$ tokens with the preceding keys and values cached,
\begin{equation}
 c_{F,k}(s)=A_{F,k}+B_{F,k}p+H_{F,k}.
 \label{eq:prediction-flops}
\end{equation}
The saved prompt length and native-token position give $p$ separately for
each model. The model ordering is unchanged throughout the measured panels;
FLOP concentration sorts the resulting per-token values, including their
prefix dependence. Qwen3-30B-A3B uses eight active experts out of 128 per layer,
so only those eight feed-forward networks enter $I_\ell^{\rm act}$.

For a request with $m\ge1$ input tokens and $n\ge1$ generated tokens,
causal prefill processes $m(m+1)/2$ attention pairs. Only its final position
requires vocabulary logits for generation. The remaining $n-1$ predictions
use cached decoding at prefix lengths $m+1,\ldots,m+n-1$. Thus
\begin{equation}
 \mathcal F_k(m,n)=
 mA_k+\frac{B_km(m+1)}{2}+H_k
 +\sum_{j=1}^{n-1}\left[A_k+B_k(m+j)+H_k\right].
 \label{eq:request-flops}
\end{equation}
This function increases with requested output length. Reserving its value
at a request's output cap therefore bounds the counted FLOPs of every
completion admitted under that cap. Model configurations, their source
revisions, and file hashes accompany the FLOP-count implementation.

\subsection{What the bound means}
\label{app:bound}

\textbf{Local witness bound.}
Fix a prefix and a target token. For a candidate model class $\mathcal C$
containing the tested family, let $d_{\mathcal C}(t)$ be the infimum FLOP count at that prefix of
a member whose deterministic prediction equals that token. Whenever a tested
member succeeds, $d_{\mathcal C}(t)\le d_F(t)$. The comparison fixes the model class, prediction rule, and supplied prefix.

\textbf{Exact-reference selector.}
Assume all panel models share a token alphabet, decode greedily with a fixed
deterministic tie-breaking rule, and have additive next-token FLOP costs $c_k(s)>0$, available prefix caches,
and no switching or state-construction overhead. Suppose every token of a finite reference $y$
is resolved. A selector may choose any model at each prefix, with advance
knowledge of the full reference and all panel predictions. Write $\operatorname{rollout}(\sigma,x)$ for the response generated by a
deterministic selector $\sigma$ with prompt $x$, and $C(\sigma)$ for its
additive prediction cost. Among selectors that reproduce $y$ exactly,
\begin{equation}
 \min_{\sigma:\,\operatorname{rollout}(\sigma,x)=y} C(\sigma)
 =\sum_{t=1}^{T}d_F(t).
 \label{eq:oracle}
\end{equation}
To attain the right-hand side, choose a cheapest matching model at position
one. Inductively, the generated prefix then equals the reference prefix, so
the cheapest matching model at the next position emits the required token.
Conversely, every selector reproducing $y$ must choose a matching model at
each position and therefore incurs at least $d_F(t)$ there. Summing proves
Eq.~\ref{eq:oracle}. If stopping is to be reproduced, it must also be included
in the reference alphabet and coverage assumption.

Equation~\ref{eq:oracle} characterizes selectors constrained to one reference.
For a checker-passing $y$, the task-correct objective
$\min_{\sigma:V(x,\operatorname{rollout}(\sigma,x))=1}C(\sigma)$ allows all such
selectors and those producing other correct responses. Its optimum is therefore
at most $\sum_t d_F(t)$ under the same assumptions. This establishes the
direction of the bound when alternative solutions are allowed.
The empirical corpus includes unresolved tokens and separately tokenized
families; the cost proxy omits map acquisition and model-state transfer.
This exact-reference result interprets the measured quantity under its
stated assumptions, while the allocation studies evaluate concrete policies.

\section{Additional reproduction results}
\label{app:robustness}
\begin{figure}[t]
\centering
\includegraphics[width=0.94\textwidth]{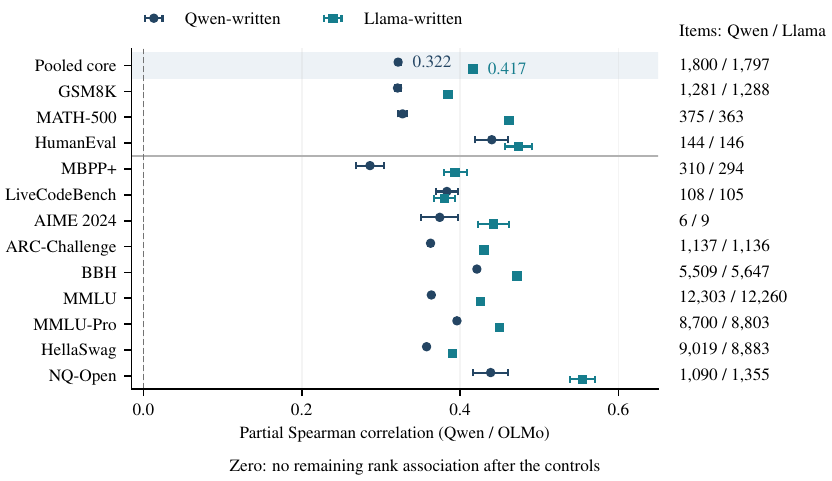}
\caption{Cross-family agreement across benchmark subsets.
Partial Spearman correlation of Qwen/OLMo word indices after controlling
generator surprisal, probe surprisal, and position; bars show $\pm$ one
item-bootstrap SD. Zero denotes no remaining rank association, and larger
positive values indicate stronger shared ordering. Pooling uses the three
core datasets. The right column gives each generator's verified item count
(Qwen/Llama), including six/nine AIME items; Table~\ref{tab:datasets}
specifies the evaluated populations.}
\label{fig:agreement}
\end{figure}

The results below supplement the concentration and shared-ordering
findings. Figure~\ref{fig:agreement} restores the per-dataset view summarized
by Table~\ref{tab:agreement-summary}; Table~\ref{tab:dataset-correlations}
gives its numerical estimates. Tables~\ref{tab:robustness}
provide additional controls.

\begin{table}[!t]
\centering\small
\caption{Per-dataset Qwen/OLMo partial correlations with bootstrap SDs (standard errors).}
\label{tab:dataset-correlations}
\begin{tabular}{lcc}
\toprule
Dataset & Qwen text & Llama text \\
\midrule
GSM8K & $0.321 \pm 0.004$ & $0.385 \pm 0.003$ \\
MATH-500 & $0.327 \pm 0.006$ & $0.462 \pm 0.005$ \\
HumanEval & $0.440 \pm 0.021$ & $0.474 \pm 0.017$ \\
LiveCodeBench subset & $0.383 \pm 0.014$ & $0.381 \pm 0.013$ \\
NQ-Open & $0.439 \pm 0.022$ & $0.555 \pm 0.016$ \\
MBPP+ & $0.286 \pm 0.018$ & $0.394 \pm 0.015$ \\
AIME 2024 & $0.374 \pm 0.023$ & $0.442 \pm 0.020$ \\
ARC-Challenge & $0.363 \pm 0.003$ & $0.430 \pm 0.002$ \\
BBH & $0.421 \pm 0.002$ & $0.472 \pm 0.001$ \\
MMLU-Pro & $0.396 \pm 0.001$ & $0.450 \pm 0.001$ \\
MMLU & $0.364 \pm 0.001$ & $0.426 \pm 0.001$ \\
HellaSwag & $0.358 \pm 0.001$ & $0.391 \pm 0.001$ \\
\bottomrule
\end{tabular}
\end{table}

\begin{table}[!t]
\centering\footnotesize
\caption{Sensitivity of cross-family agreement, with bootstrap SDs (standard errors). The frequency/entropy and loss-gap rows extend the primary controls successively. Other rows specify separate controls or populations.}
\label{tab:robustness}
\begin{tabular}{lcc}
\toprule
Variant & Qwen text & Llama text \\
\midrule
Primary controls & $0.322 \pm 0.003$ & $0.417 \pm 0.003$ \\
+ word length and token counts & $0.321 \pm 0.003$ & $0.416 \pm 0.003$ \\
+ frequency and approximate entropy & $0.299 \pm 0.003$ & $0.406 \pm 0.003$ \\
+ both families' loss gaps & $0.276 \pm 0.003$ & $0.409 \pm 0.003$ \\
+ smallest-model surprisals & $0.322 \pm 0.003$ & $0.409 \pm 0.003$ \\
High-confidence reference words & $0.287 \pm 0.004$ & $0.400 \pm 0.003$ \\
Top-five criterion & $0.302 \pm 0.008$ & $0.443 \pm 0.006$ \\
$p\ge0.5$ criterion & $0.355 \pm 0.003$ & $0.451 \pm 0.003$ \\
Second verified response & $0.328 \pm 0.003$ & $0.409 \pm 0.003$ \\
Memorization-screened subset & $0.321 \pm 0.003$ & $0.416 \pm 0.003$ \\
GSM-Symbolic & $0.319 \pm 0.002$ & $0.384 \pm 0.002$ \\
\bottomrule
\end{tabular}
\end{table}


\begin{table}[!t]
\centering\small
\caption{Gini coefficients of architecture-derived prediction FLOPs among resolved tokens; parentheses give unresolved-token percentages among all tokens. These values exclude the imputed costs used in Figure~\ref{fig:concentration}.}
\label{tab:gini}
\begin{tabular}{llccc}
\toprule
Text & Panel & GSM8K & MATH-500 & HumanEval \\
\midrule
Qwen & Qwen & 0.547 (0.76\%) & 0.451 (0.49\%) & 0.514 (0.14\%) \\
Qwen & Olmo & 0.330 (2.80\%) & 0.365 (2.99\%) & 0.316 (2.08\%) \\
Llama & Qwen & 0.543 (4.15\%) & 0.547 (4.76\%) & 0.243 (0.73\%) \\
Llama & Olmo & 0.385 (5.29\%) & 0.467 (8.72\%) & 0.218 (1.42\%) \\
\bottomrule
\end{tabular}
\end{table}

\subsection{Concentration with unresolved-token costs}
\label{app:concentration-conventions}

Figure~\ref{fig:concentration} uses the same 1,800 Qwen-generated verified
references across the Qwen, OLMo, and R1-Distill-Qwen panels: 1,281 GSM8K,
375 MATH-500, and 144 HumanEval responses. Each panel uses its own native
tokenization. The distribution bars divide by all reference tokens,
including unresolved positions.

An unresolved position has no observed successful agent and therefore no
finite measured sufficient compute under Definition~\ref{def:sufficient}.
For the displayed cumulative-cost curves, we assign it twice the
architecture-derived prediction FLOPs of the panel's largest model at
that prefix. Writing $k_{\max}$ for this model, the plotted cost is
\begin{equation}
 \widetilde d_F(t)=
 \begin{cases}
 d_F(t), & \mathcal A_F(t)\ne\varnothing,\\
 2c_{F,k_{\max}}(s_t^\star), & \mathcal A_F(t)=\varnothing.
 \end{cases}
 \label{eq:imputed-concentration}
\end{equation}
The multiplier is an assumed cost for the visualization. It does not
establish what an untested model would cost or that an unresolved token
necessarily requires a larger model. The matching outcomes themselves
are unchanged by this convention.

Figure~\ref{fig:concentration} assigns each unresolved position twice the cost of the largest panel model. This section reports the same shares among resolved tokens only (top 10\%: 51--60\% for Qwen2.5, 40--44\% for OLMo-2, 26--32\% for R1-Distill-Qwen) and Gini coefficients of resolved costs (Tables~\ref{tab:gini} and~\ref{tab:concentration-conventions}); allocation is uneven under both conventions.

Within each panel and dataset, tokens are ordered by decreasing plotted
cost. The top-$p$ share divides their cumulative cost by the total cost
of all reference tokens. Fractional percentile boundaries are evaluated
by linear interpolation of cumulative token counts and costs.
Table~\ref{tab:concentration-conventions} gives the values displayed at
1\% and 10\%, alongside the corresponding resolved-only calculation.
In the resolved-only calculation, both the percentile and total cost use
resolved tokens. The two columns therefore summarize different populations.

\begin{table}[t]
\centering\small
\caption{Concentration on Qwen-generated verified references under two cost conventions. Unresolved is the percentage of all native tokens with no matching panel model. Resolved-only columns give FLOP shares for the top 1\% / top 10\% of resolved tokens. All-token columns give the corresponding shares among all tokens after assigning each unresolved position twice the largest model's prediction FLOPs. All entries are percentages.}
\label{tab:concentration-conventions}
\begin{tabular}{@{}llrrrr@{}}
\toprule
Panel & Dataset & Unresolved & Resolved only & All tokens  \\
\midrule
Qwen & GSM8K & 0.76 & 41.76 / 60.11 & 56.86 / 79.54\\
Qwen & MATH-500 & 0.49 & 35.84 / 50.85 & 60.14 / 71.98  \\
Qwen & HumanEval & 0.14 & 45.18 / 56.72 & 52.94 / 63.66  \\
OLMo & GSM8K & 2.80 & 14.09 / 41.09 & 17.22 / 68.57  \\
OLMo & MATH-500 & 2.99 & 15.15 / 44.39 & 16.25 / 69.37  \\
OLMo & HumanEval & 2.08 & 14.84 / 39.55 & 19.87 / 63.75 \\
R1 distill & GSM8K & 5.73 & 11.84 / 31.67 & 11.47 / 74.76 \\
R1 distill & MATH-500 & 4.77 & 11.68 / 26.03 & 13.27 / 71.01 \\
R1 distill & HumanEval & 3.59 & 9.91 / 31.59 & 15.02 / 67.19 \\
\bottomrule
\end{tabular}
\end{table}

The resolved-only Gini coefficients in Table~\ref{tab:gini} use the measured
architecture-derived costs without imputation. They also include
Llama-generated references, which provide a comparison of reference sources.
The pooled resolved-only Qwen cost share of 3.03\% for half the assigned
FLOPs is another such summary; it should not be substituted for the
all-token percentile shares in Figure~\ref{fig:concentration}.

\subsection{Controls and reference variation}
\label{app:controls}

\textbf{Control variants.}
Table~\ref{tab:robustness} reports each variant with its own bootstrap SD. Frequency is estimated from the available text, and entropy is
approximate as described above. The loss-gap controls use each family's
smallest and largest model on the same reference. The high-confidence subset
requires every assigned generator token to have stored probability at least
0.5; it retains 306,490 Qwen-text words and 290,328 Llama-text words. This is
stricter than merely asking whether a token is the generator's argmax.
Surface-feature regressions on ranks, using digit/operator indicators,
word length, frequency, and position, have $R^2=0.025$ and $0.032$.
These regressions quantify that feature set's linear explanatory power.

\textbf{Reference variation and memorization screening.}
The second-target analysis uses the next available checker-passing sample
among the original four candidates. Its pooled correlations are 0.328 and
0.409; mean item demand correlates 0.574 and 0.653 across two targets from
the same generator. A separate screening campaign tests answer recall
without demonstrated solution ability and continuation of problem text.
Screening covers six model sizes: Qwen 0.5, 1.5, 7, and 72B,
and OLMo 7 and 32B.
The rule flags 33 of the 1,758 intersection items. Removing flagged identifiers
from the analysis leaves correlations near the originals; the pooled 0.5B
handover difference at $\tau=0.25$ changes from $-0.504$ to $-0.500$.
This checks sensitivity to these flags, without certifying that
unflagged items were absent from training.


\textbf{Difficulty and generator extensions.}
The MATH-500 difficulty stratification gives Qwen/OLMo correlations of
0.315 and 0.460 on levels 1 to 3, versus 0.337 and 0.464 on levels 4 and 5,
for Qwen and Llama text. The harder groups contain 177 and 166 verified
items, respectively. On Qwen3-30B-A3B text, the pooled core correlation is
$0.520$ across 293,082 words. The R1-Distill-Qwen panel
correlates 0.255/0.379 with Qwen and 0.267/0.369 with OLMo, on Qwen/Llama
text.

\textbf{Acceptance proxies.}
For each of Qwen and OLMo, we compare its smallest and largest model using
greedy disagreement and a soft overlap computed from cached truncated
probabilities. Cross-family partial correlations are 0.150/0.199 for greedy
rejection and 0.179/0.384 for soft rejection, versus 0.322/0.417 for
reproduction indices. Truncated overlap omits probability mass and is a
proxy for acceptance, not an exact full-vocabulary speculative acceptance
probability. These correlations compare the operational signals under the
same controls.


\textbf{Scope of cross-family comparisons.}
The cross-family comparisons do not separately control architecture,
tokenization, training data, or instruction tuning. The R1-Distill-Qwen
panel shares Qwen lineage, and independently developed families may also
share training data. Positive associations therefore establish shared
ordering within the evaluated panels, without identifying its source or
demonstrating that a map transfers between families. The reference-variation
and memorization-screening analyses test sensitivity to particular
alternatives; they do not establish the absence of training overlap or
memorization.

\subsection{Comparison with FlexiDepth allocations}
\label{app:flexi}
\begin{table}[!t]
\centering\small
\caption{Agreement with FlexiDepth layer allocations, with bootstrap SDs. Each split has 100 responses; the first word is excluded.}
\label{tab:flexi}
\begin{tabular}{lrcc}
\toprule
Task & Words & Qwen panel & OLMo panel \\
\midrule
copy & 24,770 & $0.061 \pm 0.008$ & $0.062 \pm 0.008$ \\
continue & 7,755 & $0.277 \pm 0.049$ & $0.267 \pm 0.048$ \\
summarize & 3,840 & $0.283 \pm 0.014$ & $0.294 \pm 0.013$ \\
repeat & 3,864 & $0.130 \pm 0.005$ & $0.120 \pm 0.004$ \\
add & 5,121 & $0.289 \pm 0.013$ & $0.291 \pm 0.013$ \\
product & 7,533 & $0.405 \pm 0.011$ & $0.389 \pm 0.012$ \\
\bottomrule
\end{tabular}
\end{table}

The released FlexiPatterns records comprise 600 generations and 82,783 labeled
native tokens. We align layer allocations and panel indices to words and
omit the first word of each response because it includes prefill effects.
The pooled analysis subtracts each task split's mean from allocation and
panel index, then computes partial rank correlation controlling log position
and word length. Within-split estimates use the same controls. This centering
procedure removes split means before the rank transformation. The pooled
partial correlations are 0.209 for Qwen and 0.213 for OLMo.
On the 68 verified addition outputs,
the Qwen correlation is $0.273$. Multiplication provides an
allocation comparison on incorrect generations, with no verified subset.

\section{Intervention protocols and full results}
\label{app:handover-details}

\textbf{Evaluation population and conditions.}
The handover experiments use the 1,758 core problems with verified references
from both generators (1,275 GSM8K, 347 MATH-500, and 136 HumanEval).
The supplied prefix is taken from the Qwen reference. A smaller model and
the 72B control start from the same prefix; subsequent tokens are generated
without additional reference tokens. Accuracy differences are paired by
problem after averaging 16 completions per condition.
The length sweep uses 0.5B, 1.5B, and 7B at
$L\in\{1,2,4,8,16,32,64\}$ from the reference midpoint.
Full-suffix handovers additionally include 32B and vary
$\tau\in\{0,0.25,0.5,0.75,0.9\}$.

\textbf{Prefix construction and pairing.}
For a reference with $C$ characters, a handover at fraction $\tau$ starts at
$\lfloor\tau C\rfloor$, snapped backward to whitespace. The generated text
is appended to this supplied prefix before verification. All models receive
the same problem and prefix within a comparison. Each condition has 16
completions at temperature 0.7 and top-$p$ 0.95. Seeds are deterministic
functions of item and position for full-suffix handovers, or item and seed
index for the length sweep. Reusing seeds organizes paired conditions; it
does not couple model outputs token by token.

The full-suffix analysis contains 35,156 paired item/size/start cells.
Four of the nominal $1,758\times4\times5$ cells collapse because snapped
start positions coincide. The pooled size ordering has zero inversions among
15 adjacent comparisons. The starting-position ordering has two inversions
among 16 comparisons. Table~\ref{tab:suffix} supplies all dataset-specific estimates.

\textbf{Length-sweep control.}
The smaller model emits at most $L$ native tokens, possibly stopping earlier.
Its sampled text is appended to the verified midpoint prefix; 72B then
receives that combined text in a new request and continues. The control is
a single uninterrupted 72B request from the midpoint. Both 72B requests
have the usual dataset output cap (512 for GSM8K and HumanEval, 1,024 for
MATH-500). Thus the intervention can generate up to $L$ additional tokens
before the 72B continuation and resets the sampler at the handback.
There are $1,758\times7\times3\times16=590,688$ small-model
completion conditions, each compared with its item/length control.

\begin{table}[!t]
\centering\footnotesize
\caption{Handover-length accuracy differences (percentage points), with bootstrap SDs (standard errors).}
\label{tab:length}
\begin{tabular}{rccc}
\toprule
Maximum tokens & 0.5B & 1.5B & 7B \\
\midrule
1 & $-0.09 \pm 0.19$ & $-0.30 \pm 0.16$ & $-0.06 \pm 0.15$ \\
2 & $-0.17 \pm 0.22$ & $-0.27 \pm 0.18$ & $-0.04 \pm 0.16$ \\
4 & $-0.30 \pm 0.24$ & $-0.17 \pm 0.19$ & $-0.00 \pm 0.18$ \\
8 & $-0.86 \pm 0.26$ & $-0.71 \pm 0.21$ & $-0.18 \pm 0.19$ \\
16 & $-2.19 \pm 0.30$ & $-1.37 \pm 0.26$ & $-0.22 \pm 0.21$ \\
32 & $-4.53 \pm 0.39$ & $-2.43 \pm 0.30$ & $-0.55 \pm 0.23$ \\
64 & $-9.44 \pm 0.50$ & $-4.68 \pm 0.39$ & $-1.50 \pm 0.30$ \\
\bottomrule
\end{tabular}
\end{table}

\begin{table}[!t]
\centering\small
\caption{Full-suffix handover accuracy differences (percentage points). Prefix fractions use characters, snapped to a word boundary.}
\label{tab:suffix}
\begin{tabular}{lrrrrr}
\toprule
Dataset & $\tau$ & 0.5B & 1.5B & 7B & 32B \\
\midrule
GSM8K & 0.0 & -52.88 & -23.39 & -4.51 & -0.16 \\
GSM8K & 0.25 & -48.25 & -21.17 & -4.12 & -0.50 \\
GSM8K & 0.5 & -35.08 & -13.92 & -3.17 & -0.74 \\
GSM8K & 0.75 & -20.60 & -7.15 & -1.77 & -0.34 \\
GSM8K & 0.9 & -10.62 & -3.07 & -0.72 & -0.08 \\
MATH-500 & 0.0 & -55.76 & -28.85 & -5.08 & +1.17 \\
MATH-500 & 0.25 & -57.01 & -28.91 & -5.49 & -0.76 \\
MATH-500 & 0.5 & -45.59 & -20.55 & -4.11 & -0.05 \\
MATH-500 & 0.75 & -32.17 & -13.08 & -3.19 & -0.32 \\
MATH-500 & 0.9 & -16.35 & -7.53 & -1.49 & +0.02 \\
HumanEval & 0.0 & -60.98 & -33.04 & -6.71 & -2.80 \\
HumanEval & 0.25 & -53.91 & -29.50 & -10.02 & -2.11 \\
HumanEval & 0.5 & -41.36 & -19.39 & -6.20 & -1.61 \\
HumanEval & 0.75 & -40.26 & -22.10 & -6.43 & -1.93 \\
HumanEval & 0.9 & -18.56 & -7.59 & -2.27 & +0.93 \\
\bottomrule
\end{tabular}
\end{table}

\textbf{Reference cost and continuation loss.}
Across the same 1,758 items, the fraction of suffix words assigned to the
14B rung or higher has Spearman correlation 0.195 with the midpoint 0.5B
accuracy loss. This aggregate reference-cost statistic provides only a
weak item-level predictor of the outcome of the handover. A handover
changes the generating policy and hence subsequent prefix states, so its
effect includes deviations, alternative solution paths, and recovery.
This comparison separates the local matching measurement from the
downstream consequences of choosing a policy for a longer continuation.

\section{Allocation protocols and complete policy comparisons}
\label{app:allocation}

This appendix specifies the full-dataset routing and drafting
experiments in Figure~\ref{fig:allocation}, including the timing projection.

\subsection{Model routing and speculative decoding on MATH-500}
\label{app:allocation-additional}
\label{app:allocation-live}

The allocation experiment in Figure~\ref{fig:allocation} evaluates ten
policies on all 500 MATH-500 problems. Table~\ref{tab:allocation-projected}
reports every point in that figure and additionally logical FLOPs.
These policies use Qwen2.5-0.5B-Instruct as the small model and
Qwen2.5-72B-Instruct as the target, with greedy decoding and a common
1,024-token output cap. Each policy's answer is checked independently using
the same symbolic-equivalence checker. No problem is excluded because the
target or another policy gives an incorrect answer.

\begin{table}[t]
\centering\small
\caption{All ten allocation policies on the full MATH-500 evaluation set (500 problems). Accuracy is independently checked for each generated response. Projected time includes prompt processing and cached model calls under Appendix~\ref{app:latency-projection}; it excludes oracle-map acquisition and is not measured deployment latency. Values are means $\pm$ one item-bootstrap SD. Confidence thresholds control small-model selection; adaptive drafting stops below probability 0.4 with a 16-token cap. MoA-map policies use advance reference-token agreement.}
\label{tab:allocation-projected}
\begin{tabular}{@{}lrrr@{}}

\toprule

Policy & Accuracy (\%) & Projected time (s) & Compute (TFLOPs) \\

\midrule

0.5B only & $28.20 \pm 2.04$ & $1.84 \pm 0.04$ & $0.67 \pm 0.01$ \\

72B only & $69.20 \pm 2.11$ & $19.03 \pm 0.38$ & $101.92 \pm 1.87$ \\

\midrule

Confidence routing, $\theta=0.5$ & $40.00 \pm 2.25$ & $2.80 \pm 0.06$ & $86.57 \pm 1.90$ \\

Confidence routing, $\theta=0.8$ & $60.00 \pm 2.21$ & $4.45 \pm 0.10$ & $99.50 \pm 1.85$ \\

Confidence routing, $\theta=0.95$ & $66.60 \pm 2.21$ & $5.91 \pm 0.13$ & $99.16 \pm 1.84$ \\

Confidence routing, $\theta=0.99$ & $68.40 \pm 2.14$ & $7.59 \pm 0.18$ & $100.38 \pm 1.89$ \\

MoA-map routing & $69.60 \pm 2.10$ & $5.12 \pm 0.26$ & $96.71 \pm 1.86$ \\

\midrule

Fixed drafting, 16 tokens & $69.00 \pm 2.13$ & $4.98 \pm 0.11$ & $154.30 \pm 3.27$ \\

Adaptive drafting & $69.00 \pm 2.13$ & $4.69 \pm 0.10$ & $140.52 \pm 2.75$ \\

MoA-map drafting & $69.40 \pm 2.11$ & $3.99 \pm 0.09$ & $112.07 \pm 2.31$ \\

\bottomrule

\end{tabular}
\end{table}

\paragraph{Information supplied to the oracle.}
For each problem, a fresh greedy target response provides a reference
$z=(z_1,\ldots,z_T)$, including its recorded termination token where present.
The map stores
\begin{equation}
 b_t=\mathbf{1}\!\left[
 \arg\max_v\pi_{0.5\mathrm{B}}(v\mid x,z_{<t})=z_t\right].
 \label{eq:live-map}
\end{equation}
This is the small-model matching component of the sufficient-compute
measurement. The target response is retained regardless of correctness,
so the map describes agreement with that response rather than a guarantee
of a correct answer. The controller receives the matching indicators. Reference generation and
map acquisition are supplied information and are excluded from policy
resource counts. Thus the experiment isolates the allocation opportunity
available with advance knowledge of agreement.

\paragraph{Routing policies.}
The confidence baseline implements the TokenCascade Chow rule
\citep{narasimhan2025faster}. At the current generated prefix $s$, it uses
the small model when $\max_v\pi_{0.5\mathrm{B}}(v\mid s)\ge\theta$ and
otherwise requests a target prediction. We evaluate all four thresholds
$\theta\in\{0.5,0.8,0.95,0.99\}$, together with standalone small- and
target-model baselines. MoA-map routing selects the small model at mapped
positions with $b_t=1$. Before emitting that prediction, it checks that the
actual small-model token equals the stored reference token; otherwise it
requests the target. After the generated prefix departs from the reference,
it uses the target. All routing policies use the same batch scheduler,
which collects deferred target requests after at most 16 small-model steps.
The target processes any missing prefix positions when it is next invoked.

\paragraph{Speculative policies.}
The fixed baseline proposes up to 16 small-model tokens before target
verification. The adaptive baseline follows confidence-based dynamic
lookahead \citep{mamou2024dynamicblog}: it appends each proposed token and
stops the block if that token's probability is below 0.4, a termination
token is reached, or 16 tokens have been proposed. MoA-map drafting uses
the length of the upcoming consecutive run of $b_t=1$, capped at 16 and by
the remaining output allowance. A zero-length run invokes the target
directly. After departure from the mapped prefix, it uses the adaptive
baseline. For every policy, the target verifies a proposed block through
greedy token equality, accepts its matching prefix, and supplies the first
correction token or a bonus token when the full block matches. Both caches
are truncated to the resulting accepted prefix.

\paragraph{Execution and resource accounting.}
The live allocation runs use a common PyTorch 2.9.1 research backend with
bfloat16 weights, math SDPA attention, and float32 vocabulary projection.
The target is distributed over four A100 80GB GPUs; the draft shares one
of these GPUs. Batches contain at most 32 problems, and a forward call
processes at most 128 new positions. This backend differs from the vLLM
reference-generation pipeline used for the reproduction study.

The logical FLOP ledger applies the architecture terms in
Appendix~\ref{app:flops} to actual forwarded positions and requested logits.
It includes prompt prefill, cache catch-up, rejected drafts, and target
verification. The ledger does not
measures memory traffic or communication.

\paragraph{Complete comparisons and their interpretation.}
Against confidence routing at $\theta=0.99$, MoA-map routing increases
accuracy from 68.4\% to 69.6\% and reduces projected time from 7.59 to
5.12 seconds.
MoA-map drafting proposes 301,691 tokens across the dataset, compared with
447,731 for the fixed-window policy, a 32.6\% reduction. Accepted draft-token
counts are similar: 268,577 and 269,286. The map therefore mainly reduces
proposals that do not survive verification. The mean number of target calls
is also similar, 57.62 versus 56.87 per response. Projected times are
3.99 versus 4.98 seconds.
Table~\ref{tab:allocation-paired} gives paired comparisons for every tested
routing threshold and both speculative baselines.

\begin{table}[t]
\centering\small
\caption{Paired differences between each MoA-map policy and the indicated baseline on the same 500 problems (map minus baseline). Positive accuracy differences favor the map; negative time differences favor the map. Accuracy is in percentage points (pp), and time differences are in seconds. Values show the mean difference $\pm$ one item-bootstrap SD; projected time follows Appendix~\ref{app:latency-projection}.}
\label{tab:allocation-paired}
\begin{tabular}{@{}lrrr@{}}
\toprule
Baseline & $\Delta$ accuracy (pp) & $\Delta$ projected (s) \\
\midrule
Routing, $\theta=0.5$ & $29.60 \pm 2.16$ & $2.32 \pm 0.24$  \\
Routing, $\theta=0.8$ & $9.60 \pm 1.60$ & $0.67 \pm 0.23$ \\
Routing, $\theta=0.95$ & $3.00 \pm 1.04$ & $-0.79 \pm 0.22$ \\
Routing, $\theta=0.99$ & $1.20 \pm 0.62$ & $-2.46 \pm 0.21$ \\
Fixed drafting & $0.40 \pm 0.29$ & $-1.00 \pm 0.04$ \\
Adaptive drafting & $0.40 \pm 0.29$ & $-0.71 \pm 0.03$  \\
\bottomrule
\end{tabular}
\end{table}

\paragraph{Output agreement and uncertainty.}
Finite-precision block and incremental evaluations need not produce identical
greedy predictions. Map fallback occurs on 100 routing
responses and 169 speculative responses. All departures and length-capped
responses remain in the accuracy and resource totals. Consequently, small
accuracy differences should be read as empirical outcomes of this backend,
rather than evidence that changing speculative draft length intrinsically
improves an exactly preserved target policy.

Means and paired differences use 2,000 bootstrap resamples of the 500 problem
identifiers. Tables report one bootstrap SD, which describes sampling
variation across problems rather than repeated timing runs. 
Projection uncertainty from the timing assumptions is examined separately
in Appendix~\ref{app:latency-projection}.


\subsection{Projection from published inference benchmarks}
\label{app:latency-projection}

We separate the measured answer accuracies from an illustrative latency
projection. Accuracies use all 500 MATH-500 problems generated with our
experimental backend. The horizontal coordinates instead describe a warm,
single-request execution scenario anchored to the official Qwen2.5 BF16
benchmarks for vLLM~0.6.3 on A100~80GB GPUs: four GPUs for the 72B target and
one for the 0.5B model \citep{qwen25speed}.
We assume both models remain resident and execute sequentially without
interference. The published one-token-input rates are 31.37 and 311.55 output
tokens per second, respectively, for 2,048-token outputs. These rates measure
whole requests, rather than isolated cached forwards.

For each model $m$, let $K_m$ denote the number of recorded forward calls and
$F_m$ the total number of input positions processed, including prompt tokens,
cache catch-up, and rejected speculative tokens. We use
\begin{equation}
\widehat T=\sum_{m\in\{\mathrm{small},\mathrm{target}\}}
 \left[a_m K_m+b_m(F_m-K_m)\right].
\end{equation}
A single-position call costs $a_m$; processing $n$ positions together costs
$a_m+b_m(n-1)$. Thus, parallel verification need not cost as much as $n$
sequential target calls. Prompt processing is included in the same ledger,
so it is not added a second time. The projection retains the recorded call
counts, including chunked processing, and is computed separately for every
response before averaging across items.

We anchor $a_m$ at the inverse of the shortest-input published generation
rate. To set $b_m$, write the published total time at input length $P$ as
$T_m(P)=2048/v_m(P)$ and fit
\begin{equation}
 b_m=\mathop{\arg\min}_{b\geq0}
 \sum_{P\in\{6144,14336,30720\}}
 \left[T_m(P)-2048a_m-b(P-1)\right]^2.
\end{equation}
This gives $(a_m,b_m)=(3.210,0.03683)$ milliseconds for 0.5B and
$(31.878,0.28557)$ milliseconds for 72B. The fit reproduces these published
whole-request times with a maximum relative error of 2.13\%.
The fit describes the published whole-request measurements; cached block-verification timings are not separately calibrated.

The substantive assumption is transferring the marginal cost of a longer
input to extra positions within a cached forward call. Longer-input timings
also include increased attention cost during subsequent generation;
therefore, the public benchmark does not separately identify prefill,
cache catch-up, or verification costs. The projection does not model their
kernel differences, context-dependent nonlinearities, model-switching
costs, or additional controller overhead. It also excludes queueing,
model loading, network delivery and oracle-map acquisition. Consequently,
these coordinates are conditional model-execution estimates, not measured
end-to-end deployment latencies or a demonstrated vLLM speedup.
The answer accuracies remain those of the original backend; identical
outputs after migration to another engine have not been established.

We assess sensitivity using bulk-position coefficients scaled by
$\{0,0.5,1,2\}$, small-model times scaled by $\{0.5,1,2\}$, and an added
cost of $\{0,0.5,1\}$ target-call equivalents per speculative verification.
All 36 combinations are applied to the full dataset, without selecting
parameters according to policy ranking. Within these scenarios, MoA-map
drafting has lower projected latency than fixed-window and adaptive
drafting; its relative reductions range from 8.3--24.7\% and 7.3--18.4\%,
respectively. These ranges describe chosen assumptions, not statistical
intervals or guaranteed prediction bounds. In particular, they do not
cover arbitrary policy-specific controller overhead or unobserved block
cost nonlinearities.

\section{Relation to earlier measurements}
\label{app:related-full}

Sequence prediction has long been formulated through policies and future
rewards \citep{ranzato2016sequence,bahdanau2017actor}. Controlled Decoding
and Transfer $Q^*$ use this formulation for decoding-time alignment
\citep{mudgal2024controlled,chakraborty2024transfer}.
COLLAB selects off-the-shelf model policies in a shared decoding MDP using
long-term utility \citep{chakraborty2025collab}; CITER formulates
small/large-model token routing as an MDP \citep{zheng2025citer}.
Our measurement assigns a cost to a
specified next-token action at a reference state. The handover and allocation
experiments then examine the consequences of letting selected agents generate
the subsequent states. This connects the measurement to the policy-switching
problem while keeping local reproduction and terminal task reward distinct.

Several prior methods expose token-dependent computational behavior through
a particular inference system. Duo-LLM searches internal module allocations
to minimize perplexity under a compute budget \citep{alizadeh2024duollm}.
FS-GEN examines disagreement and collaboration frequencies for small and
large models \citep{zhang2024fsgen}. FlexiDepth records a learned layer
allocation policy within Llama-3-8B \citep{luo2025flexidepth}. Our contribution
is the cross-family measurement design: independently score a shared,
checker-passing trajectory across several model sizes, compare the resulting
orderings, and test their relation to continuation outcomes.

This design retains the exact-reference criterion's dependence on wording.
Judge Decoding demonstrates that valid continuations can be rejected by
alignment-based acceptance \citep{bachmann2025judge}. Top-five and probability
thresholds change our matching rule but still evaluate a chosen reference.
The shared-prefix experiment therefore measures reproducibility of a
successful trajectory, while the handovers assess task correctness after
the trajectory is allowed to change.

Generation-stage switching is an established design class. Reasoning Relay
studies the transfer of partial reasoning between models
\citep{lu2025relay}; LLM Shepherding uses a larger model's prefix to support
a smaller model \citep{dong2026shepherding}; RelayGen selects segment-level
switch points for efficient inference \citep{song2026relaygen}.
The handover experiments here vary capacity, starting position, and duration
under a common task checker. They characterize the relationship between
conditional reproduction and sustained continuation, rather than introducing
model switching itself.

\end{document}